\documentclass[runningheads]{llncs}
\usepackage{graphicx}
\usepackage{acronym}
\usepackage{amsfonts} 
\usepackage{amsmath}
\usepackage{subcaption}
\usepackage{hyperref}
\usepackage{comment}
\acrodef{IF}[IF]{Isolation Forest}
\acrodef{LOF}[LOF]{Local Outlier Factor}
\acrodef{XAI}[XAI]{eXplainable Artificial Intelligence}
\acrodef{AD}[AD]{Anomaly Detection}
\acrodef{DSSs}[DSSs]{Decision Support Systems}
\acrodef{AcME}[AcME]{Accelerated Model-agnostic Explanations}
\acrodef{NDCG}[NDCG]{Normalized Discounted Cumulative Gain}
\usepackage{xspace}
\usepackage[colorinlistoftodos, textwidth=2.5cm]{todonotes}
\PassOptionsToPackage{xcolor}{dvipsnames}
\newcommand{\ourmethod}{AcME-AD\xspace}

\newcommand{\chiara}[1]{\textcolor{orange}{#1}}

\newcommand{\vale}[1]{\textcolor{cyan}{#1}}

\begin{document}
\title{\ourmethod: Accelerated Model Explanations for Anomaly Detection}
%
%
\author{Valentina Zaccaria\thanks{First and second author equally contributed.}\inst{1}\orcidID{0009-0008-7930-1434} \and David Dandolo\inst{2}\orcidID{0000-0001-9875-0931} \and
Chiara Masiero\inst{2}\orcidID{0000-0003-1948-049X} \and 
Gian Antonio Susto \inst{1} \orcidID{0000-0001-5739-9639}}
\authorrunning{V. Zaccaria et al.}
%
\institute{Department of Information Engineering, University of Padova\\via G. Gradenigo 6/b, Padova (Italy)\\
\email{\{valentina.zaccaria,gianantonio.susto\}@unipd.it}
\and
Statwolf Data Science\\via della Croce Rossa 42, Padova (Italy) 
\email{\{david.dandolo,chiara.masiero\}@statwolf.com\thanks{This work was partially carried out within the MICS (Made in Italy – Circular and Sustainable) Extended Partnership and received funding from Next-GenerationEU (Italian PNRR – M4C2, Invest 1.3 – D.D. 1551.11-10-2022, PE00000004). Moreover this study was also partially carried out within the PNRR research activities of the consortium iNEST (Interconnected North-Est Innovation Ecosystem) funded by the European Union Next-GenerationEU (Piano Nazionale di Ripresa e Resilienza (PNRR) – Missione 4 Componente 2, Investimento 1.5 – D.D. 1058 23/06/2022, ECS00000043). This work was also cofunded by the European Union in the context of the Horizon Europe project ‘AIMS5.0 - Artificial Intelligence in Manufacturing leading to Sustainability and Industry5.0’ Grant agreement ID: 101112089.}}
}
\maketitle              
\begin{abstract}
Pursuing fast and robust interpretability in Anomaly Detection is crucial, especially due to its significance in practical applications. Traditional Anomaly Detection methods excel in outlier identification but are often `black-boxes', providing scant insights into their decision-making processes. This lack of transparency compromises their reliability and hampers their adoption in scenarios where comprehending the reasons behind anomaly detection is vital. At the same time, getting explanations quickly is paramount in practical scenarios. To bridge this gap, we present AcME-AD, a novel approach rooted in Explainable Artificial Intelligence principles, designed to clarify Anomaly Detection models for tabular data. \ourmethod transcends the constraints of model-specific or resource-heavy explainability techniques by delivering a model-agnostic, efficient solution for interpretability. It offers local feature importance scores and a what-if analysis tool, shedding light on the factors contributing to each anomaly, thus aiding root cause analysis and decision-making. This paper elucidates \ourmethod's foundation, its benefits over existing methods, and validates its effectiveness with tests on both synthetic and real datasets. AcME-AD's implementation and experiment replication code is accessible in a public repository\footnote{\url{https://github.com/dandolodavid/ACME/tree/master/notebook/anomaly_detection_notebook}}.
\keywords{Anomaly Detection  \and  Explainable Artificial Intelligence   \and Interpretable Machine Learning \and Outlier Detection \and Unsupervised Learning.}

\end{abstract}

\section{Introduction}
\label{sec:Introduction}

\ac{AD} is a machine learning task that aims at identifying data objects that significantly deviate from a normal behavior observed in the majority of the data \cite{hawkins1980identification}. \ac{AD} approaches are typically unsupervised: this is particularly appealing given that in many scenarios the annotation process is both expensive and error-prone. \ac{AD} applications span various domains, including intrusion detection, fraud detection in banking and finance \cite{awoyemi2017credit}, healthcare \cite{ukil2016iot}, manufacturing \cite{susto2017anomaly} and fault detection in industrial processes and equipment \cite{barbariol2020self,brito2022explainable}. 


Many \ac{AD} approaches have been presented in the literature, for example ensembles (e.g., tree-ensembles like Isolation Forest \cite{liu2008isolation}) or neural network-based approaches like autoencoders \cite{zhou2017anomaly}; these algorithms identify anomalous points by providing a real-valued score, called \textit{anomaly score}, quantifying the level of outlyingness of the specific data point. Typically, such approaches can be considered `black-boxes' given their non-interpretable nature, lacking the capability to offer direct explanations or interpretations for their decisions.

Providing explanations for identified anomalies, for example by indicating how the input features contribute to the predicted anomaly score (i.e., \textit{feature attribution}), can be exceptionally beneficial. In real-world scenarios, one of the main obstacles in the adoption of \ac{AD} models is the absence of a performance evaluation procedure, caused by the inherent unsupervised nature of the task. Consequently, shedding light on the inner workings of models can boost trust and confidence in them and foster their employment.
Additionally, feature attribution for individual predictions (\textit{local explanations}) is a valuable tool, for example in industrial settings, to conduct \textit{root cause analysis}. This means identifying the reason for abnormal behavior and possibly the solutions to fix it, reducing the effort from domain experts and operators/users involved in monitoring and troubleshooting.

While research interest in \ac{XAI} has been increasing, it has focused primarily on supervised machine learning approaches such as regression and classification \cite{saeed2023explainable}. Only recently the research community has pointed its attention toward explainability for \ac{AD} models \cite{li2023survey} and, in general, to unsupervised task \cite{peng2022xai}.

In this work we propose \ourmethod, a feature attribution approach based on \ac{AcME} \cite{dandolo2023acme}, specifically designed to explain unsupervised \ac{AD} models for tabular data. 

\ac{XAI} in the context of \ac{AD} poses unique challenges. Our method addresses all the main ones:
\begin{description}
    \item[Flexibility] In \ac{AD}, as extensively demonstrated in \cite{han2022adbench}, there is no unique method that outperforms the others for every dataset. This fact restricts the application of model-specific yet effective explainability techniques, such as those proposed for \ac{IF} in \cite{carletti2023interpretable} or for Principal Component Analysis in \cite{takeishi2019shapley}.  
    Instead, \ourmethod operates with any AD model, requiring only an anomaly score and a threshold to define outliers. This allows users to leverage the best-performing model while still obtaining interpretable results.
    \item[Computational Efficiency] Often \ac{AD} algorithms are employed in critical scenarios, for instance, intrusion detection or fault detection, where timing is essential. Therefore, all the methods that are based on SHAP \cite{lundberg2017unified}, even on its most efficient model-agnostic version KernelSHAP, are not suitable in this scenario. Indeed, the computational burden associated with KernelSHAP becomes increasingly challenging as the dimension and number of data points grows. 
Leveraging the AcME framework, AcME-AD delivers explanations in real-time, as we demonstrate through experiments in Section \ref{sec:experiments}, making it ideal for time-critical applications where rapid decision-making is paramount.
    \item[Global vs Local Explanations] Anomalies within a dataset can be inherently diverse and not always be confined to a unique region of the input feature space. This implies that the importance of each feature in the model might differ for individual anomalous data points. 
 Recognizing the diverse nature of anomalies, \ourmethod prioritizes explanations for individual data points. This enables `what-if' scenario exploration, revealing how hypothetical feature changes affect the model's output and facilitating potential anomaly remediation strategies. In addition, \ourmethod also offers a broader understanding of the global behavior of the model, by deriving global importance scores starting from local scores of predicted anomalies.
\end{description}

In this paper, after a review of related works in Section \ref{sec:relatedworks}, we describe in detail the inner working of \ourmethod and its visualizations in Section \ref{sec:ACMEAD}. In Section \ref{sec:experiments}, we assess \ourmethod performance through experiments on both synthetic and real-world datasets. We compare it with model-specific and the state-of-the-art KernelSHAP explanations. In addition, we demonstrate its superior time efficiency with respect to KernelSHAP. Finally, we further assess the feature ranking quality using Feature Selection as a proxy task. Future work and conclusions are drawn in Section \ref{sec:conclusions}.


\section{Related Works}
\label{sec:relatedworks}

In the literature, model explainability in tabular \ac{AD} is approached from different perspectives and addressed through different techniques due to the inherent subjectivity of the task and the context dependence. In this section, we limit ourselves to the so-called \textit{feature importance} and \textit{feature attribution} methods. 
For an in-depth exploration of other approaches, like rules-based ones \cite{barbado2022rule}, we refer the interested readers to \cite{li2023survey}, which present an up-to-date and comprehensive review of the majority of state-of-the-art techniques.

Feature importance methods aim to explain model predictions of anomalies by assigning an importance score to each input feature. This score depends on the extent to which the feature has contributed to the predicted class (anomalous or normal) or to the predicted anomaly score. 
On the other hand, feature attribution methods go a step further by also illustrating the direction in which a particular feature influences the model's prediction. 

There exist some intrinsically interpretable AD models, e.g., ECOD \cite{li2022ecod}, that directly provide importance scores. However, most feature importance methods are typically \textit{post-hoc}. Post-hoc approaches are applied on the top of \ac{AD} models, after they have been trained and they are motivated by the fact that designing inherently interpretable anomaly detectors potentially requires sacrificing predictive power and accuracy, which is not ideal. Among post-hoc explanations, some \textit{model-specific} frameworks for \ac{AD} can be found in the literature. These strategies leverage the inner working of the \ac{AD} model to provide reliable and efficient explanations. For example, \cite{carletti2023interpretable,kartha2021you} illustrate two procedures to assign importance scores for Isolation Forest, both at local level, i.e., for a single data point, and global level. In \cite{arcudi2023exiffi} the authors adopt a similar approach for Extended Isolation Forest \cite{hariri2019extended}. The paper \cite{takeishi2019shapley} integrates Shapley values to Probabilistic Principal Component Analysis \cite{tipping1999probabilistic}. The work \cite{pevny2016loda} introduces LODA, an anomaly detection algorithm effective in online settings and it equips it with a post-hoc way to assign local feature importance scores. Finally, plenty of approaches in the literature are tailored for Deep Learning settings \cite{antwarg2021explaining,li2023survey}.

Unfortunately, as mentioned in Section \ref{sec:Introduction}, no single model outperforms others across all contexts, datasets and applications. This leaves the user with the option of selecting one of the aforementioned algorithms, even if it may not be the best performing one, or to giving up to explainability. In contrast, there are \textit{model-agnostic} methodologies, that can be applied to any model, offering high-portability. 
As in \ac{XAI} in general, the most used model-agnostic post-hoc methods are based on SHAP \cite{park2020explainable,brito2022explainable}. Therefore, they are feature attribution methods and they essentially show how feature perturbations modify the anomaly score. 
A well-known problem of SHAP, even in its KernelSHAP version, is its computational complexity, which makes it unsuitable to provide explanations when the \ac{AD} model is integrated into \ac{DSSs} where the goal is enabling data-driven fast decision-making. To reduce the computational time in a DSS-oriented perspective, the work in \cite{dandolo2023acme} introduced \ac{AcME}. AcME is a model-agnostic post-hoc feature attribution method for tabular data, that provides explanation both at global and local interpretability. \ac{AcME} has shown to be three orders of magnitude faster than KernelSHAP, while providing comparable explanations. However, \ac{AcME} works with regression and classification models only. In this work, we propose a modification of it, \ourmethod, specifically tailored to explain anomaly detection models for tabular data.




\section{\ourmethod: an agnostic and efficient interpretability approach for AD on tabular data}
\label{sec:ACMEAD}

This work presents \ourmethod, a novel approach designed to augment the interpretability of anomaly detection (AD) models. \ourmethod primarily focuses on local interpretability, delving into the individual features' influence on specific data point predictions. It achieves this by analyzing the impact of each feature on both the data point's anomaly score and its classification as anomalous or normal. It additionally offers a global evaluation of feature importance derived from these local explanations, providing a holistic understanding of model behavior.

Similar to \ac{AcME}, \ourmethod employs a perturbation-based approach. Local importance scores are calculated by iteratively perturbing individual feature values in an observed data point using quantiles extracted from their empirical distributions. We introduce four novel sub-scores derived from the anomaly score predicted for the perturbed data point. These sub-scores capture how modifying a feature value affects both the anomaly score and the predicted class. A weighted sum combines these sub-scores into a single local importance score, with the weights being customizable by the user. This flexibility allows tailoring explanations to specific use cases and stakeholder needs.

Finally, a pivotal aspect of \ourmethod lies in its computational efficiency when compared with state-of-the-art methods, making it suitable to be integrated in \ac{DSSs}.  This efficiency stems from its reliance on pre-computed statistics and its avoidance of costly model re-training during explanation generation.


The algorithmic procedure is outlined in Section \ref{subsec:localacmead}. In Section \ref{subsec:localvis} we discuss the local visualization that \ourmethod is equipped with - a plot inherited from \ac{AcME} that displays the predictions for both perturbed and original data points. This plot serves as a valuable tool for \textit{what-if} analysis, allowing to assess how changes in each feature would impact both the anomaly score and the classification. 
Finally, Section \ref{subsec:globalacmead} illustrates how we provide a global understanding of the model, starting from local explanation of outliers. 

\subsection{Local Interpretability}
\label{subsec:localacmead}

Consider an arbitrary \ac{AD} model with a decision function $m(\cdot)$ such that $m(\mathbf{x}) \in M \subseteq \mathbb{R}$ which assigns an anomaly score to a $p$-dimensional observation $\mathbf{x}\in {\mathbb{R}}^p$. Let $t\in \mathbb{R}$ be an hard-coded or dynamic threshold. If $m(\mathbf{x}) \leq t$ the model classifies $\mathbf{x}$ as normal; otherwise, it classifies $\mathbf{x}$ as an outlier. 

Define a mapping function $f$ that maps the anomaly score between 0 and 1 as follows:
\begin{equation}
    \begin{array}{ll}
        f: & M \rightarrow [0,1] \\
        & m(\mathbf{x}) \mapsto f(m(\mathbf{x})) = f(\mathbf{x})
    \end{array}
\end{equation}
where $f(\cdot)=1$ indicates the maximum degree of outlyingness (i.e., the maximum anomaly score), $f(\cdot)=0$ indicates the maximum degree of normality (i.e., the minimum anomaly score) and $f(\cdot)=0.5$ represents the threshold between normal and anomalous observations. 

Let $\mathbf{x}^b=\mathbf{x}$ denotes the \textit{baseline} vector, i.e., the point with respect to which perturbations and their corresponding predictions are computed. To calculate the local importance score of each feature $j \in 1,...,p$ in the prediction of $\mathbf{x}$, we perturb it based on its quantile values computed within the dataset. Meanwhile, all other features are kept fixed to their actual values. 

Specifically, we construct a \textit{variable-quantile matrix} $\mathbf{Z}_j \in \mathbb{R}^{Q \times p}$ whose rows are identical to the baseline vector, except for the \textit{j}-th components. These components are replaced by the $Q$ quantile values of the empirical distribution of the processed feature $j$. Notice that the number of rows in $\mathbf{Z}_j$ can be tuned by varying the number of selected quantiles $Q$. 
We then compute the mapped anomaly score  $f(\mathbf{x})$ of the considered observation $\mathbf{x}$ and the mapped anomalies scores $f(\mathbf{Z}_j)$, where by abuse of notation $f(\mathbf{Z}_j)$ indicates the vector of mapped anomalies scores computed for considering the rows of $\mathbf{Z}_j$.

Starting from these values, we define four novel metrics as follows: 
\begin{description}
    \item[Delta]
    \begin{equation}
        D_j \doteq \max (f(\mathbf{Z}_j)) - \min (f(\mathbf{Z}_j)) 
        \label{eq:Dj}
    \end{equation} 
    $D_j \in [0,1]$ is the difference between the maximum and minimum (mapped) anomaly score achieved by perturbing the feature $j$. A larger difference indicates a more substantial impact of the feature on the model.

     \item[Ratio]
     \begin{equation}
         R_j \doteq \frac{f(\mathbf{x}) - \min{f(\mathbf{Z}_j)}}{D_j} 
        \label{eq:Rj}
     \end{equation}
     $R_j \in [0,1]$ represents the normalized distance of the actual mapped anomaly score from the minimum attainable one through the specific feature perturbation. A normalized distance close to 1 indicates that the original feature value is already anomalous compared to other values assumed by the same feature in the dataset. Consequently, the greater the distance, the greater the importance of the feature should be. 
     
    \item[Change of predicted class]
    \begin{equation}
        C_j \doteq \begin{cases}
            1 & \text{if} \max{f(\mathbf{Z}_j)} \geq 0.5 \text{ and } \min{f(\mathbf{Z}_j)} < 0.5 \\
            0 & \text{otherwise} \\
        \end{cases}
        \label{eq:Cj}
    \end{equation}
    This score $C_j \in [0,1]$ captures the propensity of feature perturbation to induce a change in the classification of the current observation \textbf{x}, causing it to transition between anomalous and normal states. 

    The importance of $C_j$ becomes evident in practical applications of \ourmethod, for example in industrial settings or safe-critical applications. In such contexts, users are not only concerned with the increase in the anomaly score, but are also keen on understanding which specific features have triggered the alarm.

    \item[Distance to change]
    \begin{equation}
        Q_j \doteq \begin{cases}
            1-\lvert q(\mathbf{Z}_j[k,j]) - q(\mathbf{x}[j])\rvert & \text{if } C_j = 1 \\
            0 & \text{if } C_j = 0\\
        \end{cases}
        \label{eq:Qj}
    \end{equation}

    Intuitively, $Q_j$ quantifies how much the feature $j$ needs to be perturbed, in terms of quantile $q(\cdot)$, to change the classification outcome (anomalous or normal) of $\mathbf{x}$, if it can occur. The term $\mathbf{x}[j]$ is the j-th feature value assumed by observation $\mathbf{x}$. The term $\mathbf{Z}_j[k,j]$ denotes the element in position $(k,j)$ in the matrix $\mathbf{Z}_j$ that is the j-th feature perturbed value that yield class change of $\mathbf{x}$ and at the same time has the minimum distance - in terms of quantile - with respect to the original feature value $\mathbf{x}[j]$. In mathematical terms: 
    \begin{equation*}
        \begin{aligned}
            k \doteq &\arg \min_k \left\lvert q(Z_j[k,j]) - q(x[j])\right\rvert \\
            & \text{s.t.} (f(x)\leq0.5 \wedge f(Z_j[k])>0.5) \vee (f(x)\geq 0.5 \wedge f(Z_j[k])<0.5).
        \end{aligned}
    \end{equation*}
    The use of quantile levels aims to provide a relative distance that ranges from 0 to 1. The importance of the feature should increase as the required perturbation to change the class becomes smaller; hence we set $Q_j$ as 1 minus the distance. For the same reasons of $C_j$, also $Q_j$ holds practical significance.
    

\end{description}

The local importance score of feature $j$ is a convex combination of these four scores: 
\begin{equation}
    \begin{aligned}
          I_j(\mathbf{x)} & \doteq w_D D_j + w_C C_j + w_Q Q_j + w_R R_j \\
        \text{s.t.} & \; w_D + w_C + w_Q + w_R = 1, \\
      & \; w_i \geq 0, \text{for}  \; i = D, C, Q, R
    \end{aligned}
    \label{eq:Ij}
\end{equation}
The weights $w_i$ are input parameters of the algorithm, defined by the user according to the use case. Default values are set to $w_D = 0.3, w_C = 0.3, w_R =0.2, w_Q = 0.2$.

The complete procedure to compute the local importance score for a given observation is as $\mathbf{x}\in\mathbb{R}^p$ follows: 

\begin{enumerate}
    \item Set the baseline vector equal to the considered data point: 
    \begin{equation}
        \mathbf{x}^b=\mathbf{x}.
    \end{equation}
    \item For each $q\in \{0, 1/(Q-1),2/(Q-2), ..., 1\}$ create the new vector $\mathbf{z}_{j,q}\in \mathbb{R}^p$. This is obtained from $\mathbf{x}^b$ by substituting $\mathbf{x}_j^b$ with $\mathbf{x}_{j,q}^b$, i.e., the value of quantile $q$ for the j-th variable: 
    
     \begin{equation}
         \mathbf{z}_{j,q} = [x_1^b, ..., x^b_{j-1}, x_{j,q}^b,x^b_{j+1}, ..., x_p^b ].
     \end{equation}
    
    \item Create the variable quantile matrix for feature $j\in\{1,...,p\}$: 
    \begin{equation}
        \mathbf{Z}_j =
        \begin{bmatrix}
        {x}^{b}_1 & {x}^{b}_2 & \ldots & {x}^{b}_{j, 0} & \ldots & {x}^{b}_p \\
        {x}^{b}_1 & {x}^{b}_2 & \ldots & {x}^{b}_{j, 1/Q-1} & \ldots & {x}^{b}_p \\
        \vdots & \vdots & \ddots & \vdots & \ddots & \vdots \\
        {x}^{b}_1 & {x}^{b}_2 & \ldots & {x}^{b}_{j, 1} & \ldots & {x}^{b}_p \\
        \end{bmatrix}.
    \end{equation}

    \item Compute the mapped anomaly scores associated with the variable-quantile matrix rows:
    \begin{equation}
        f(\mathbf{Z_j}) = 
        \begin{bmatrix}
        f(z_{j,0}) \\
        f(z_{j,1/(Q-1)}) \\
        \vdots \\
        f(z_{j,1})
        \end{bmatrix}.
    \end{equation}

    \item Compute the four scores $D_j, R_j, C_j, Q_j$ as in Equations \ref{eq:Dj}, \ref{eq:Rj}, \ref{eq:Cj}, \ref{eq:Qj}, respectively. 

    \item Given the set of weights $w_D, w_R, w_C, w_Q \in \mathbb{R}$ such that $w_i \geq 0, \sum_i{w_i}=1,$ $i \in \{D,R,C,Q\}$, compute the local importance score as in Equation \ref{eq:Ij}.
\end{enumerate}

For categorical features, the \textit{M} distinct values that the feature could assume, are used in place of the \textit{Q} quantiles as outlined \ac{AcME} \cite{dandolo2023acme}.

\subsection{Results Visualizations}
\label{subsec:localvis}

For the visualization of local explanations, we propose two kinds of plots, the \emph{what-if} analysis tool and the Single Feature Exploration tool. 

\begin{figure}[!tb]
    \centering
    \includegraphics[width = \textwidth]{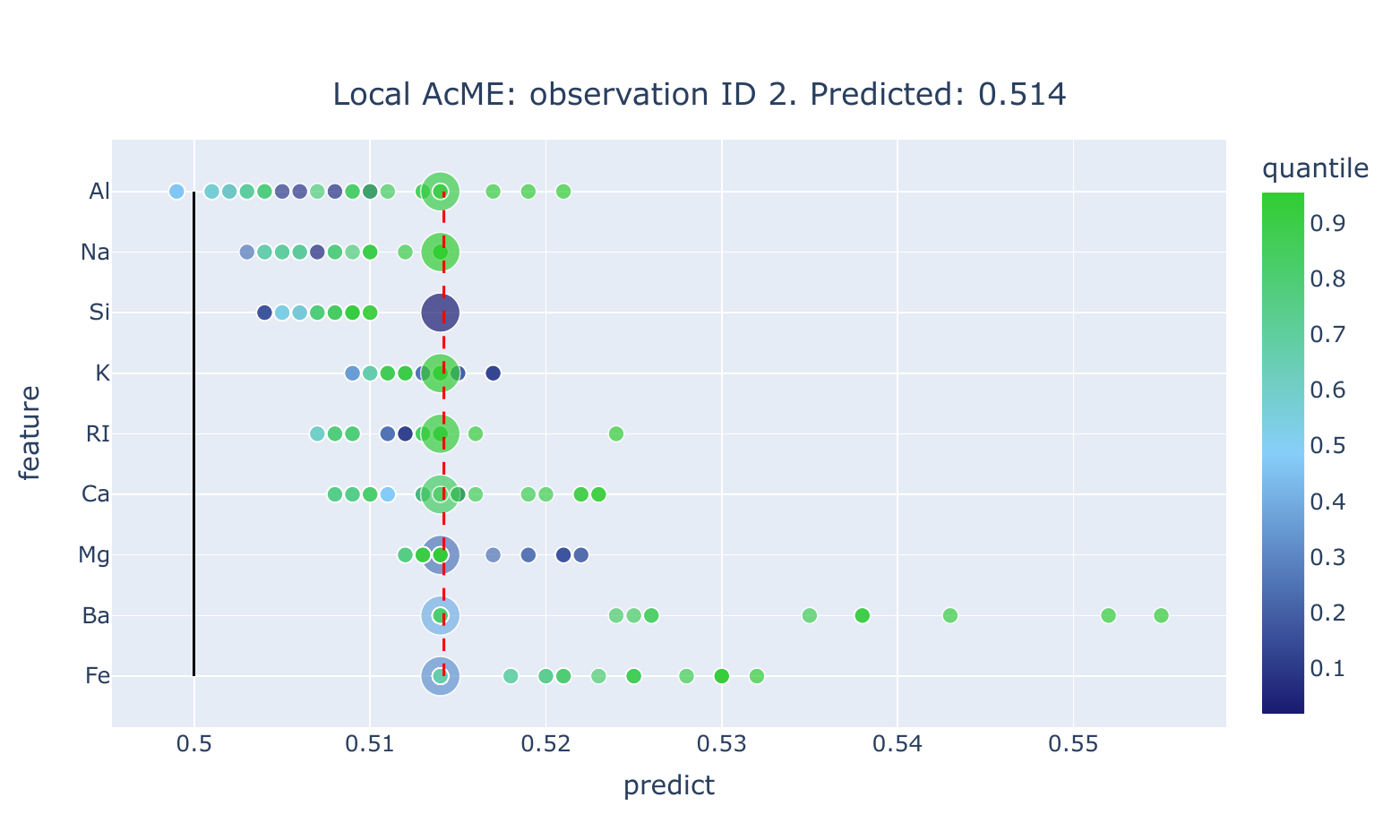}
    \caption{The \emph{what-if} analysis tool for local interpretability of prediction computed by the Isolation Forest described in Section \ref{subsec:realworlddatasets} for Sample 2 of the Glass Dataset.}
    \label{fig:loc_vis_example}
\end{figure}

The former, illustrated in Fig. \ref{fig:loc_vis_example}, is the most informative visualization. In the \emph{what-if} analysis tool, \textit{y}-axis displays the features in descending order of importance, according to Eq. \ref{eq:Ij}. On the \textit{x}-axis, the plot displays the mapped anomaly scores. For each horizontal line, corresponding to a single feature, there are Q bubbles, showing how the anomaly score of the observation changes when the corresponding feature is perturbed assuming the q-th quantile level (keeping all other features fixed).  Colors indicate the quantile level of the perturbation, ranging from the smallest quantile in blue to the greatest, in green. The vertical black line denotes the mapped anomaly score equal to 0.5, which indicates the class change (from anomalous to normal and viceversa) for the observations. This allows users to immediately discern features that can induce a class change, like \textsf{Al} in the example of Fig. \ref{fig:loc_vis_example}, where a local explanation of an anomalous observation predicted by \ac{IF} and belonging to the \texttt{Glass} dataset is displayed. Details of this experiment are presented in Section \ref{sec:experiments}. Additionally, a dashed red line corresponds to the mapped anomaly score of the original non-perturbed observation. In such a way, it is evident which features increase or decreases the anomaly score. Finally, larger circles are placed on the original anomaly score. Their colors provide direct information about the quantile in which the original feature values fall, effectively showing in which part of the empirical distribution the feature value is situated.  The \emph{what-if} analysis tool is extremely useful in practical scenarios, because it shows what features can be changed to bring an anomalous point to a normal state. For instance, in the example in Fig. \ref{fig:loc_vis_example}, perturbations of \textsf{Na} and \textsf{Si} only decrease the anomaly score, \textsf{Ba} and \textsf{Fe} increase it, while the effects of the remaining feature depend on the level of perturbation.

\begin{figure}[!tb]
    \centering
    \includegraphics[width = \textwidth]{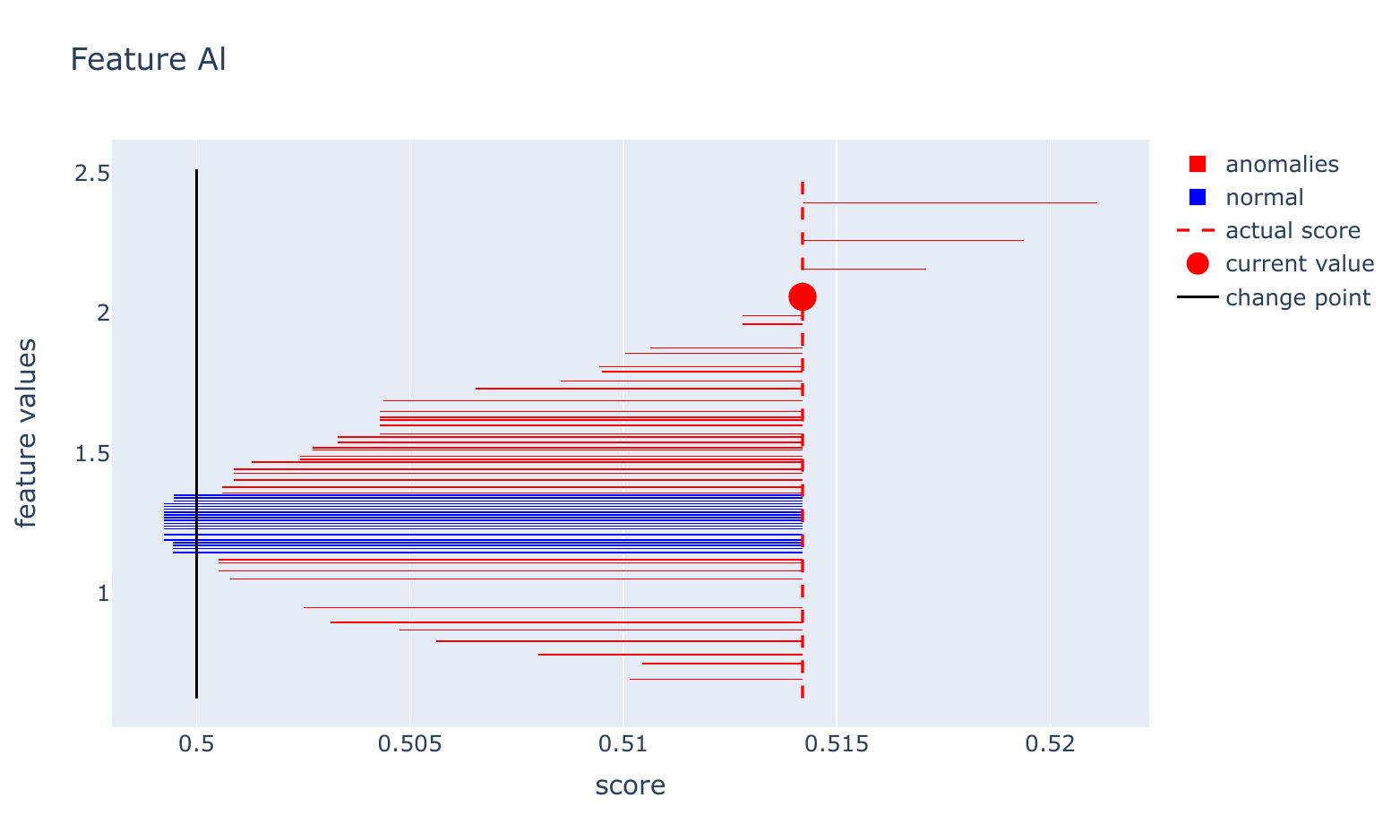}
    \caption{The Single Feature Exploration plot to interpret the prediction computed by the Isolation Forest described in Section \ref{subsec:realworlddatasets} for Sample 2 of the Glass Dataset.}
    \label{fig:feature_exp_example}
\end{figure}

An example of the Single Feature Exploration tool is illustrated in Fig. \ref{fig:feature_exp_example}. The plot selects one feature and showcases the dependence between its values and the anomaly score predicted by the model, obtained by marginalizing over the values of all the other features. Thus, this visualization resembles Partial Dependence Plots (PDPs) \cite{friedman2001greedy} and Individual Conditional Expectation (ICE) Plots \cite{goldstein2015peeking}. However, unlike PDPs, which illustrate the average effect across the dataset and ICE Plots, which display a line for each sample, the Feature Exploration plot is local as it focuses on a single observation. 
The Single Feature Exploration tool adopts a waterfall format, where the \textit{y}-axis represents distinct values assumed by the feature when perturbed according to quantiles. The \textit{x}-axis shows the anomaly score associated with each perturbation. Therefore, each bar represents the change in the anomaly score, with respect to the original anomaly score, associated with the specific feature perturbation. The original anomaly score is highlighted with a dashed red line and the red circle denotes the original feature value. A solid black line denotes the threshold at which the predicted class changes. Feature perturbations that produce an anomaly score below the threshold are represented by the color blue, the other ones by red. This visualization offers two key advantages. First, it provides an immediate understanding about how altering a specific feature value influences the anomaly score. Second, it offers insights into the decision boundary shape projected onto the feature direction. Even if the first type of plot already conveys this information by using colors, Feature Exploration plots allow for a more straightforward reading. On the other hand, since each plot is associated with one feature, as the number of features increases, the user has to analyze more plots. Having prompt decision-making in mind, we do not set this as a default visualization but prefer the above-described \emph{what-if} analysis tool to provide exhaustive local interpretation at a glance.

\newcommand{\pedfrac}[2]{\genfrac{}{}{0pt}{}{#1}{#2}} 

\subsection{Global Interpretability}
\label{subsec:globalacmead}

Let $\mathcal{S}$ be the set of interest, we defined the global importance score of feature $j$ as the sum local importance scores $I_j$ for the points in $\mathcal{S}$ predicted as anomalous:
\begin{equation}
    T_j = \sum_{\genfrac{}{}{0pt}{}{x \in \mathcal{S}}{\text{s.t.} f(x) > 0.5 }} I_j(x).
\end{equation}

\begin{figure}[!tb]
    \centering
    \includegraphics[width=\textwidth]{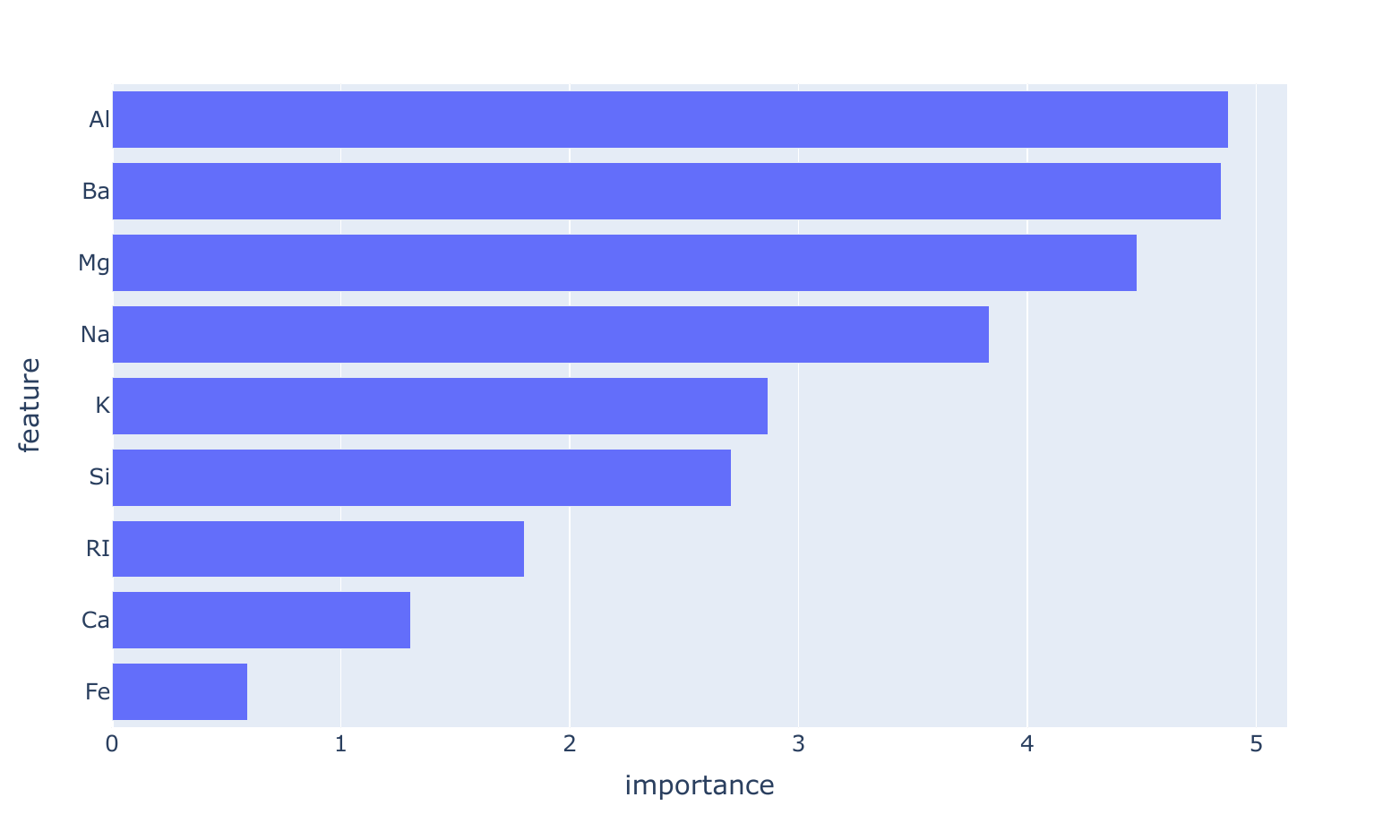}
    \caption{\ourmethod Global importance scores computed to interpret the Isolation Forest model trained on the Glass Dataset as described in Section \ref{subsec:realworlddatasets}.}
    \label{fig:global_barplot}
\end{figure}

To effectively present these scores, \ourmethod generates a bar plot like the one in Fig. \ref{fig:global_barplot}. Each feature is assigned to a bar of length equal to the global importance score. The features are organized in descending order based on their importance score. 

\section{Experimental Results}
\label{sec:experiments}
In this section, we present the experimental results for \ourmethod, evaluating its performance on both synthetic and real-world datasets.

Before showing the results, it is important to note that assessing the quality of explanations in the \ac{AD} context is challenging. While there exists an abundance of real-world datasets tailored for \ac{AD}, they often lack both labels and ground-truth explanations. 

The lack of ground-truth feature rankings and relevance scores results in the absence of a standardized evaluation process, making it difficult to evaluate the generated feature rankings. 

Furthermore, the unsupervised nature of \ac{AD} - i.e., the absence of labels - is an issue in the evaluation of an \ac{AD} model performance. This issue extends to the evaluation of explanations because, if using poorly performing models, potential misattribution of feature importance scores may be caused by the models themselves rather than reflect shortcomings in the \ac{XAI} approach. Hence, it is crucial to consider high-performing models when determining the quality of the explanations provided by the XAI approach to be evaluated. 

To overcome these issues, in Section \ref{subsec:syntheticdatasets}, we initially resort to a synthetic dataset, introduced in \cite{carletti2023interpretable}, with controlled characteristics and prior knowledge of the relevance of features. 
Then, in Section \ref{subsec:realworlddatasets}, we conduct experiments on a modified version of the well-known real-world \verb|Glass Identification| dataset originally aimed at multi-class classification. Despite its original classification purpose, which potentially hinders the performance of \ac{AD} models \cite{arcudi2023exiffi}, we select this dataset due to some prior knowledge on the most important features. 

We use \ac{IF} as \ac{AD} model, due to its high performance on the selected datasets and because it is one of the most popular shallow \ac{AD} models, thanks to its low computational cost and overall good performances \cite{han2022adbench}. 
We compare the explanations generated by \ourmethod with the explanations of KernelSHAP, the state-of-the-art model agnostic explainer in \ac{AD}. We not only consider the feature ranking in the comparison, but also the computational time, which is the key asset and the primary motivation for our method. 
Moreover, we compare \ourmethod with the model-specific feature attribution algorithm DIFFI \cite{carletti2023interpretable}. 
Finally, since in real-world applications no ground truth is available, we adopt Feature Selection to further assess the quality of generated global explanations. Feature selection is already used with this goal \cite{carletti2023interpretable,arcudi2023exiffi,pevny2016loda}, being a suitable proxy in actual unsupervised settings. The results are illustrated in Section \ref{subsec:featureselection}. To foster further research and experimentation, we have made our \ourmethod available as a Python implementation in a public repository on GitHub\footnote{\url{https://github.com/dandolodavid/ACME/tree/master/notebook/anomaly_detection_notebook}}. This repository not only provides the code for \ourmethod but also includes all the code necessary to replicate the experiments presented in this section.

\subsection{Experiments on a Synthetic Dataset}
\label{subsec:syntheticdatasets}

\subsubsection{Dataset description} 
We consider a dataset where each data point $x\in \mathbb{R}^p$ is represented by the p-dimensional vector: 
\begin{equation}
    x = [\rho \cos(\theta), \rho \sin(\theta), n_1, ..., n_{p-2}]^T
\end{equation}
where $n_j\sim\mathcal{N}(0,1)$ for $j=1,..., p-2$ are white noise samples. The parameters $\rho$ and $\theta$ are random variables drawn from continuous uniform distributions. For regular points we have: 
\begin{equation}
    \theta \sim \mathcal{U}(0,2\pi),  \quad \rho \sim \mathcal{U}(0,3) 
\end{equation}
while for anomalous points we have: 
\begin{equation}
    \theta \sim \mathcal{U}(0,2\pi),  \quad \rho \sim \mathcal{U}(4,30). 
\end{equation}

\begin{figure}
  \centering
  \begin{subfigure}{0.49\textwidth}
    \centering
    \includegraphics[width=\linewidth]{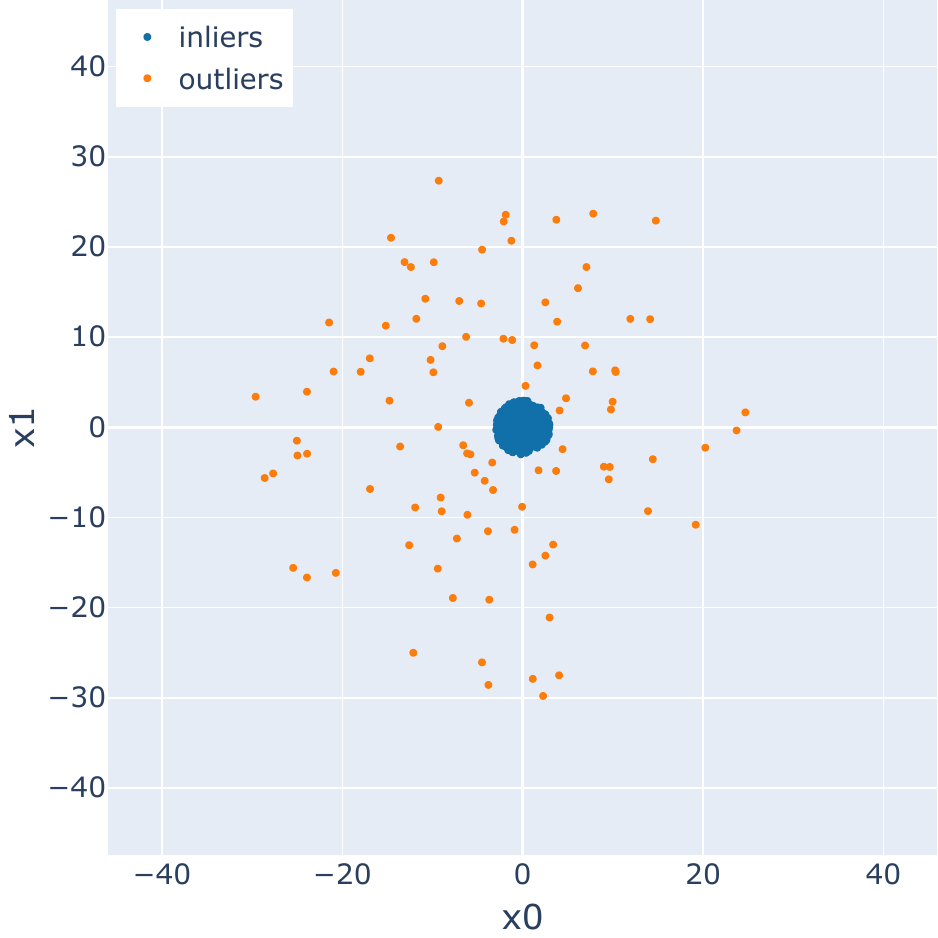}
    \caption{Training data}
    \label{fig:synthetic_train_data}
  \end{subfigure}
    \begin{subfigure}{0.49\textwidth}
    \centering
    \includegraphics[width=\linewidth]{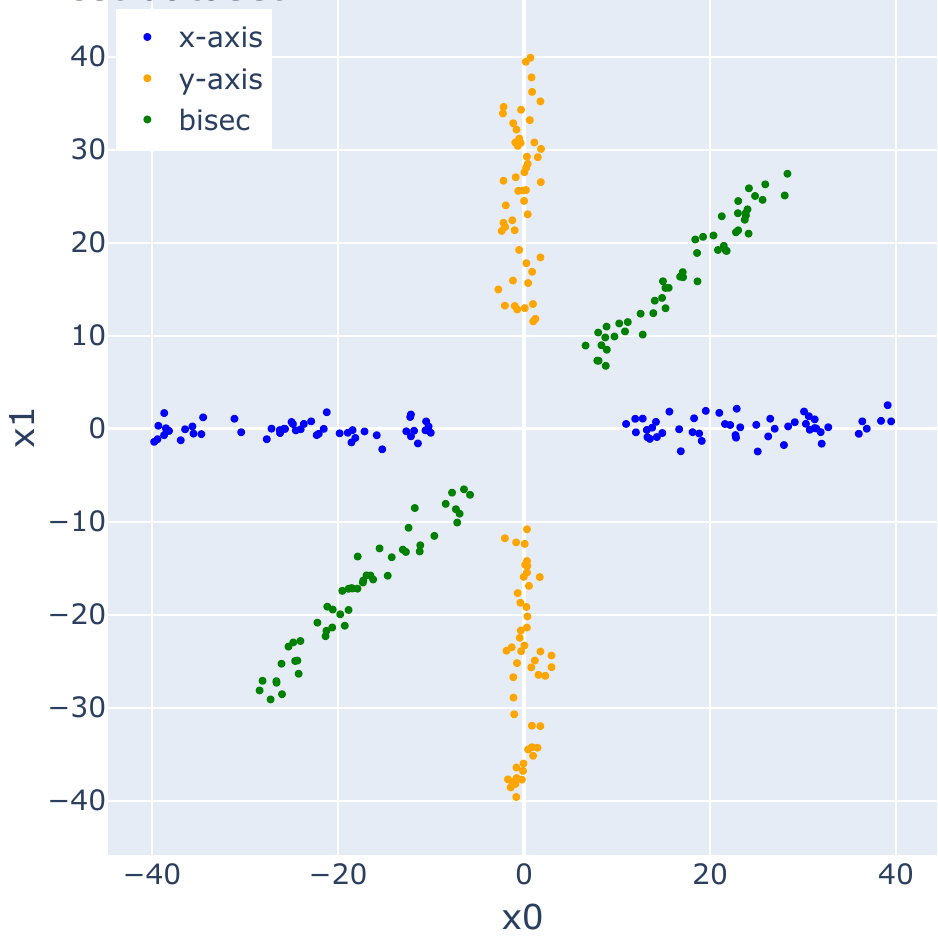}
    \caption{Test outliers}
    \label{subfig:synthetic_test_data}
  \end{subfigure}
  \caption{Synthetic datasets projected onto the first two dimensions}
  \label{fig:synthetic_data}
\end{figure}

The training set is made up of 1000 six-dimensional data points ($p=6$, $N=1000$) with a contamination factor of 10\%, i.e., containing 100 anomalous points. The first two dimensions of the training points are shown in Fig. \ref{fig:synthetic_train_data}.

For the testing phase, we generate 300 additional ad-hoc anomalies, 100 distributed along the first dimension (x-axis), 100 along the second dimension (y-axis) and 100 along the bisection line of the plane composed by the first two feature directions, as illustrated in Fig. \ref{subfig:synthetic_test_data}.
For x-axis anomalies, $x_0$ is the only relevant feature, for y-axis anomalies $x_1$ is, while for bisection anomalies $x_0$ and $x_1$ have the same relevance. 

\subsubsection{}
We train an instance of \ac{IF} with 100 trees and $\psi=256$, as suggested in the original paper \cite{liu2008isolation}, obtaining a F1-score equal to 0.7412 on training data. 
We compute predictions of the test dataset, with an F1-Score of 0.8439. 
We run \ourmethod to get the correspondence local importance scores for predicted outliers with 70 quantiles and default weights. 

\begin{figure}
  \centering

  \begin{subfigure}{0.49\textwidth}
    \centering
    \includegraphics[width=\linewidth]{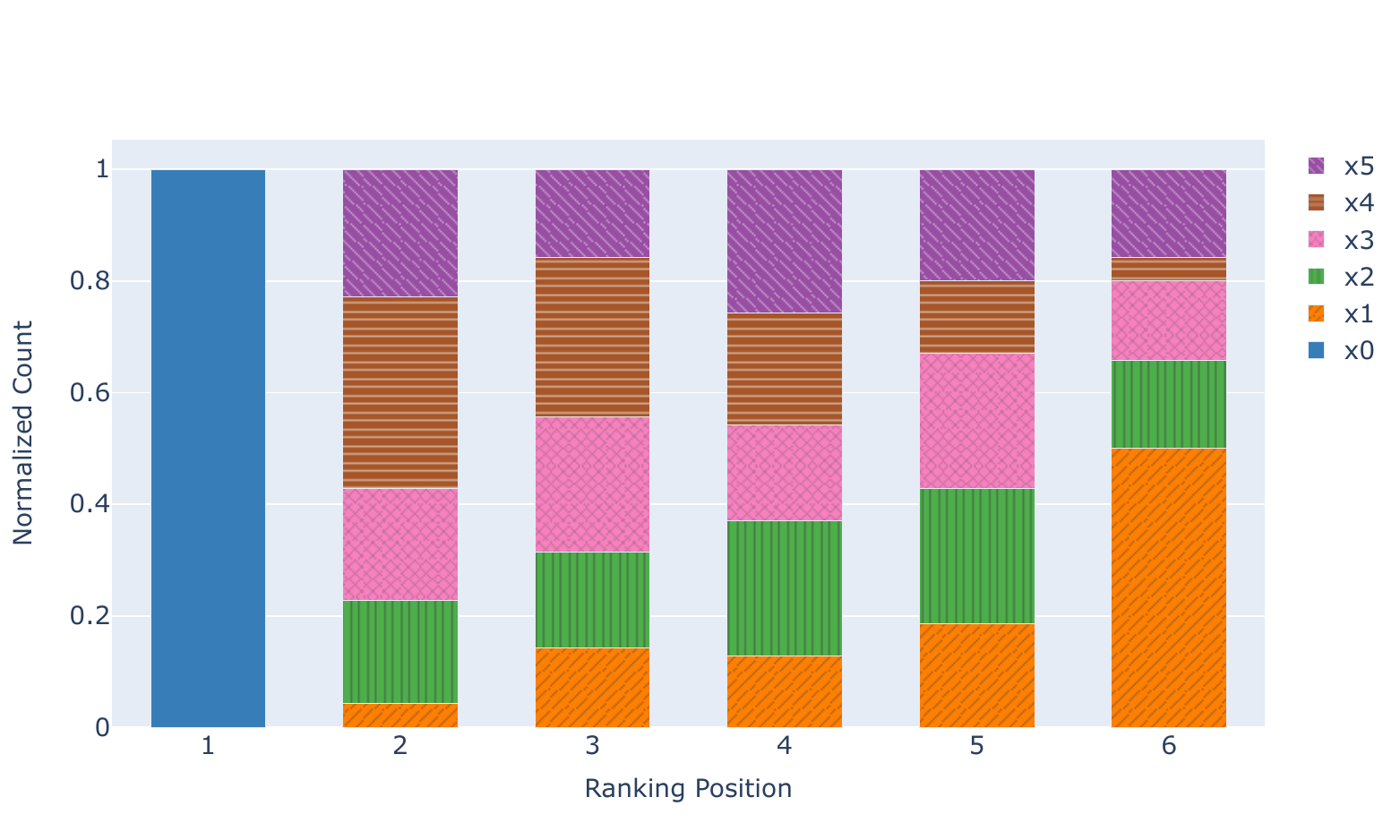}
    \label{fig:subplot1}
  \end{subfigure}
  \hfill
  \begin{subfigure}{0.49\textwidth}
    \centering
    \includegraphics[width=\linewidth]{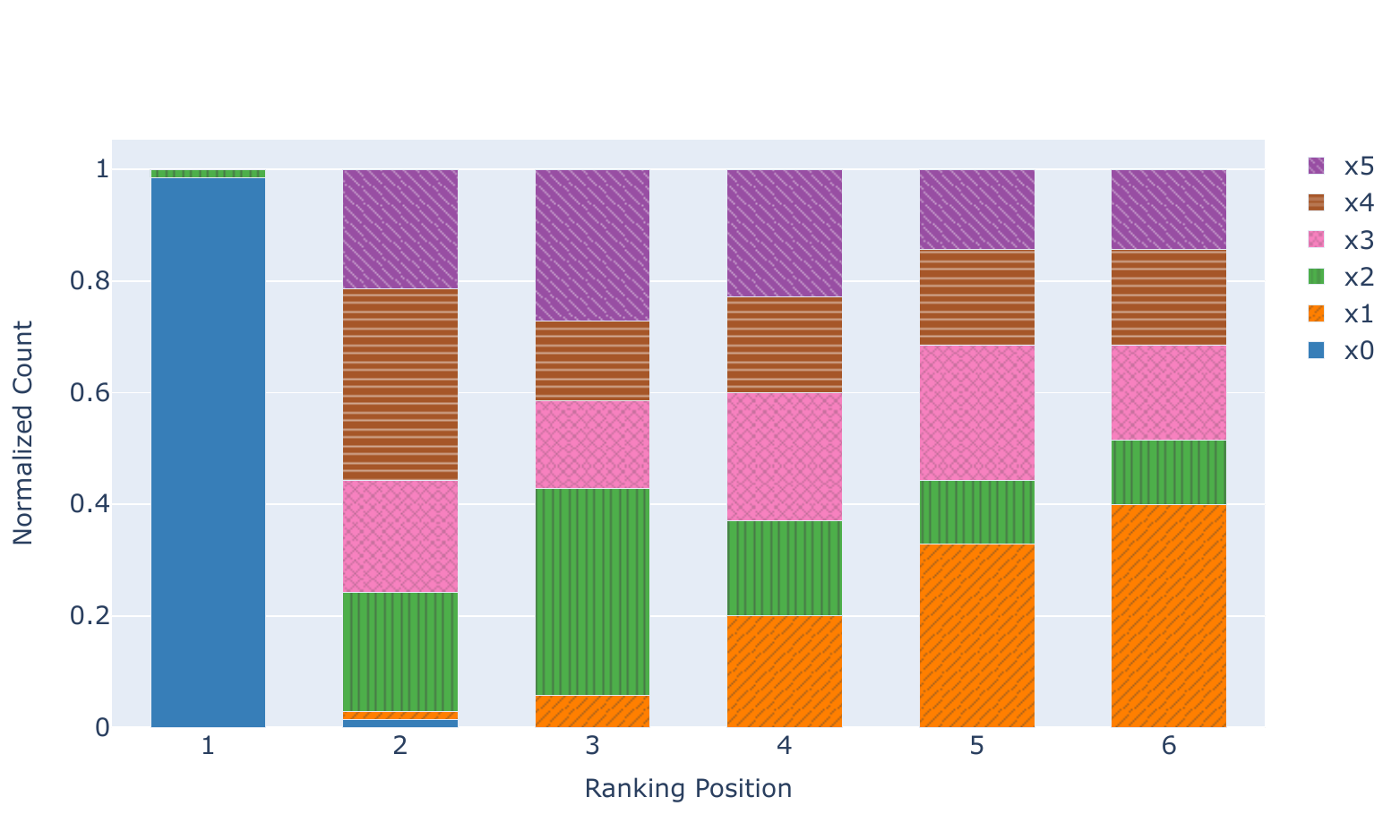}
    \label{fig:subplot2}
  \end{subfigure}
  \begin{subfigure}{0.49\textwidth}
    \centering
    \includegraphics[width=\linewidth]{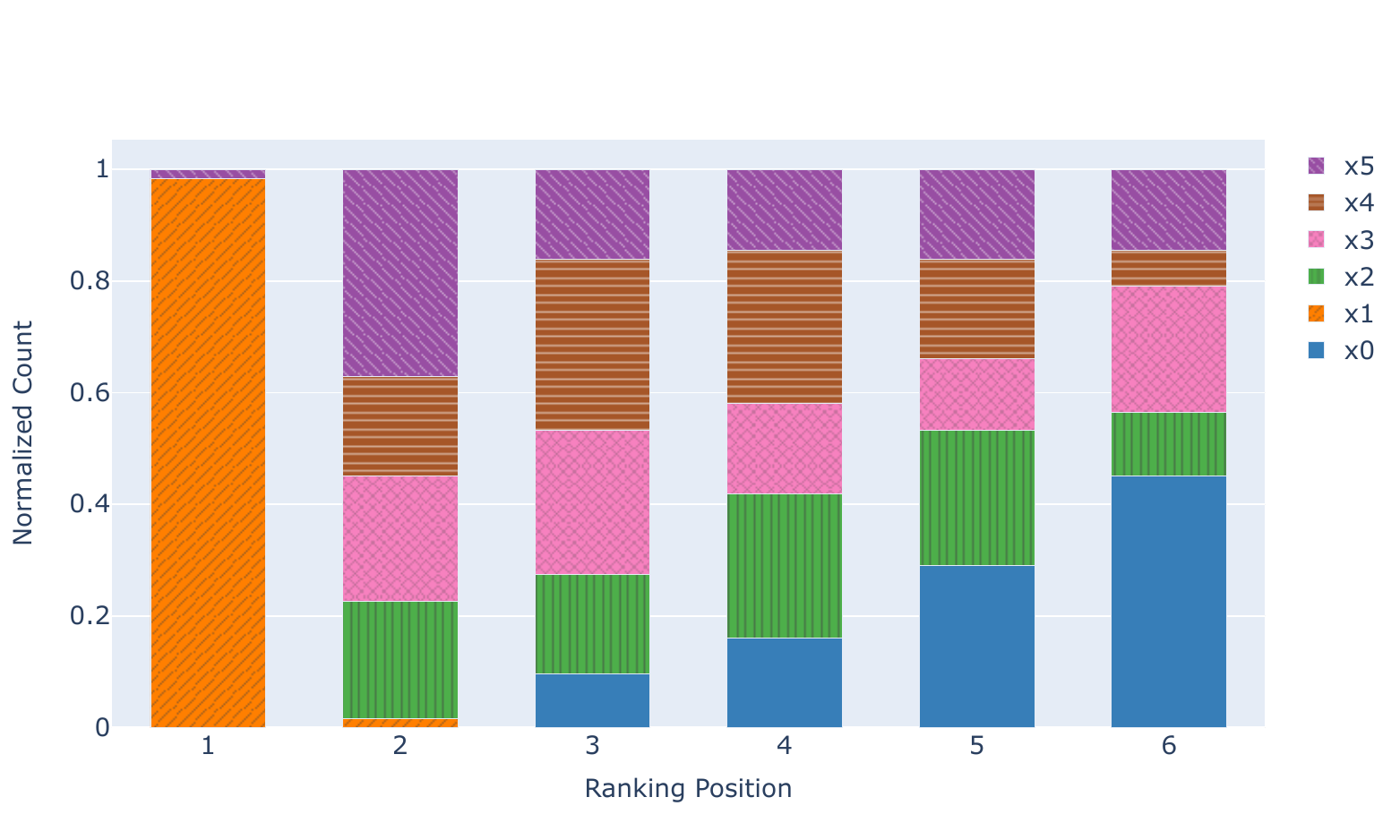}
    \label{fig:subplot3}
  \end{subfigure}
  \hfill
  \begin{subfigure}{0.49\textwidth}
    \centering
    \includegraphics[width=\linewidth]{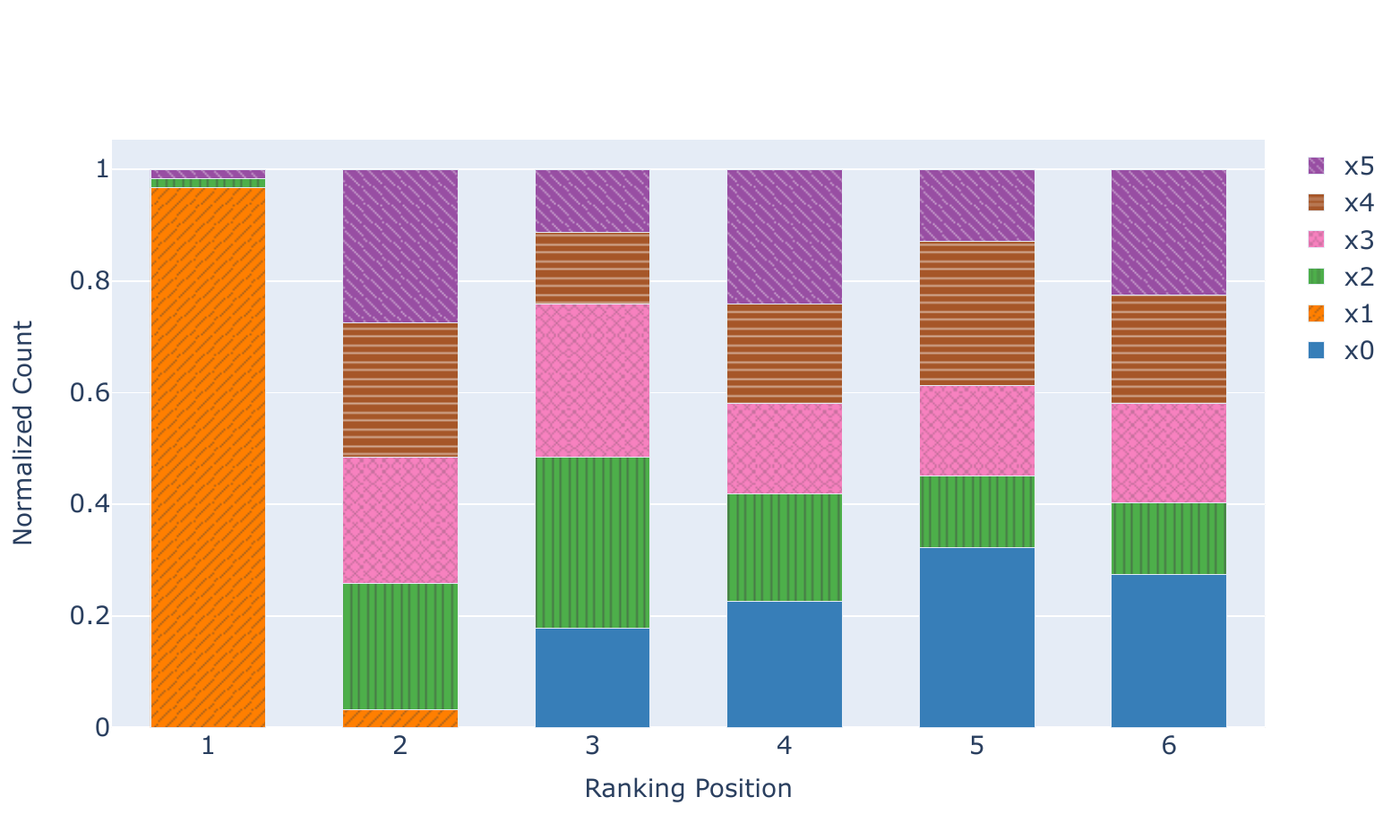}
    \label{fig:subplot4}
  \end{subfigure}
  \begin{subfigure}{0.49\textwidth}
    \centering
    \includegraphics[width=\linewidth]{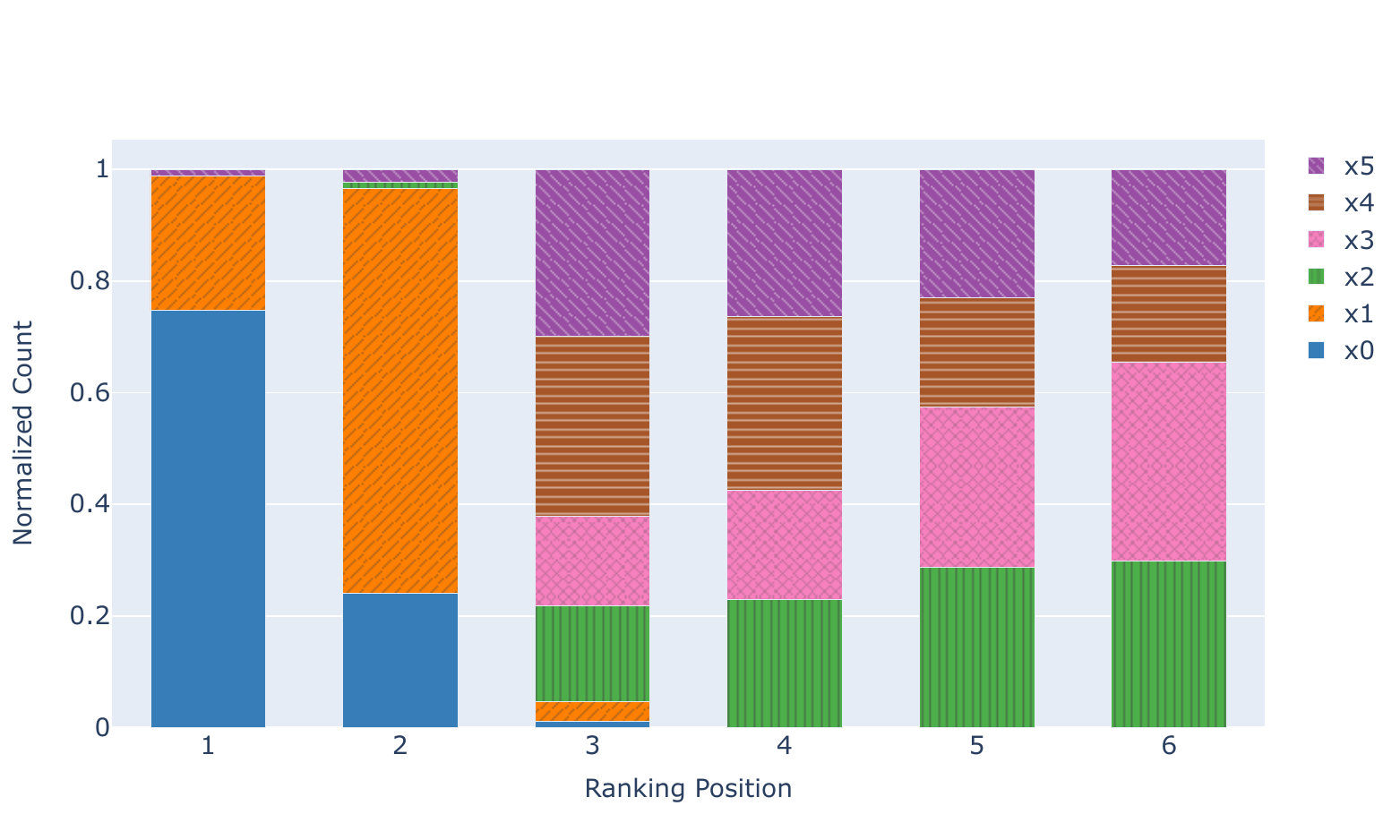}
    \caption{\ourmethod}
    \label{fig:subplot5}
  \end{subfigure}
  \hfill
  \begin{subfigure}{0.49\textwidth}
    \centering
    \includegraphics[width=\linewidth]{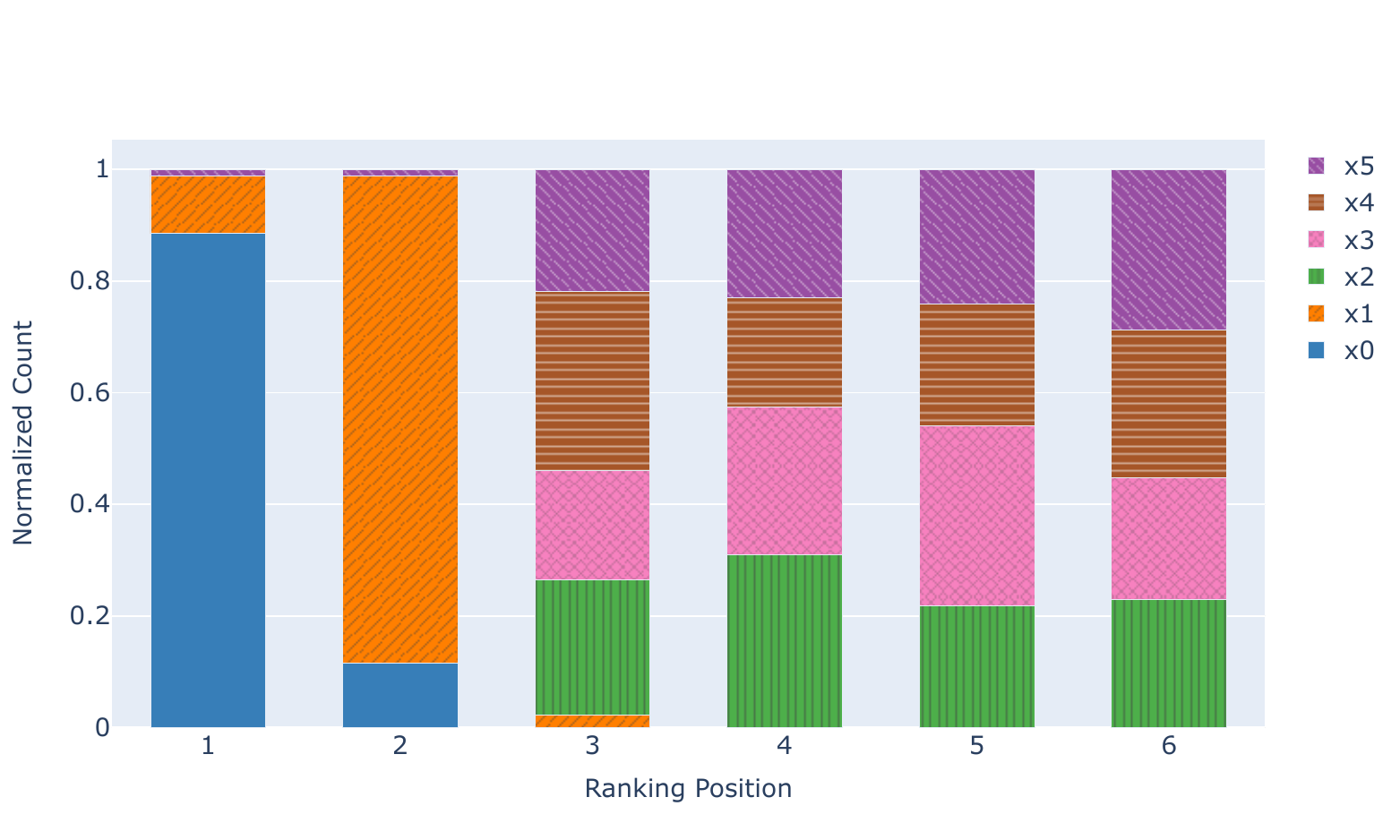}
    \caption{KernelSHAP}
    \label{fig:subplot6}
  \end{subfigure}

  \caption{Comparison of feature rankings computed by \ourmethod and KernelSHAP for an Isolation Forest trained on the synthetic dataset as detailed in \ref{subsec:syntheticdatasets}. \ourmethod in the left column, KernelSHAP in the right one. First row represents the feature ranking resulting from outliers along the x-axis, second row represents the feature ranking resulting from outliers along the y-axis, and the third rows represents the feature ranking resulting from outliers along bisec.}
  \label{fig:synthetic}
\end{figure}

For each outlier in the test set that is classified as anomalous by the model, we generate the corresponding \ourmethod local explanation. Subsequently, we categorize the local explanations based on whether the outlier belongs to the x-axis, y-axis or the bisec dataset. 
From each local explanation, we retrieve the corresponding feature ranking. We then count how many times each feature is ranked in the first position (the most important), in the second position and so on, and normalize the count.  Fig. \ref{fig:synthetic} depicts the resulting visualization, with the the normalized count for each feature ranking for x-axis, y-axis and bisec outliers in a stacked bar chart. The rankings obtained by \ourmethod and local KerneSHAP are shown respectively in the left and in the right column. It can be seen that both methods correctly identify the most important feature in the majority of observations, namely $x_0$ for the x-axis set, $x_1$ for the y-axis set and both $x_0$ and $x_1$ for the bisec dataset, aligning with prior knowledge. 

Notably, in this experiment, the average time to compute one local explanation, averaged over 100 instances, for \ourmethod is $0.170s \pm 0.033s$ and for KernelSHAP is $0.394s \pm 0.009s$. 

\subsection{Experiments on Real-world Datasets}
\label{subsec:realworlddatasets}

\subsubsection{Datasets description}
We assess the robustness of \ourmethod on two real-world datasets, \texttt{Glass Identification} and \texttt{Satellite}. 
The \texttt{Glass Identification} dataset \cite{misc_glass_identification_42}, from UCI Machine Learning Repository, was originally designed for multi-class classification, comprising seven classes corresponding to different types of glasses. The dataset contains 213 samples, each characterized by 9 numerical features: Refractive Index (\textsf{RI}) and the concentrations of Sodium (\textsf{Na}), Magnesium (\textsf{Mg}), Aluminum (\textsf{Al}), Silicon (\textsf{Si}), Potassium (\textsf{K}), Calcium (\textsf{Ca}), Barium (\textsf{Ba}) and Iron (\textsf{Fe}). 
In our analysis, we resort to a modification of the dataset, by grouping classes 1 to 4 (representing window glasses) as inliers, while classes 5 to 7 (representing non-window glasses) as outliers.
Following \cite{gupta2019beyond}, we leverage pre-existing knowledge about class 7, denoted as \textit{headlamps glass}. Specifically, we know that Aluminum and Barium should be relevant features for distinguishing headlamps glass from window glass. 

Secondly, we use the Outlier Detection DataSets (ODDS) Satellite dataset \cite{Rayana:2016}, containing 6,425 data points with 36 numerical features. This dataset's increased complexity provides a suitable benchmark to evaluate AcME-AD's computational efficiency compared to KernelSHAP, the \emph{de-facto} standard for model agnostic interpretability. 

\subsubsection{Experimental setup on Glass dataset}
{Unlike in the synthetic case, where outliers are only in the test set, in these experiments anomalies are already contained in the training dataset. In particular, they correspond to data points belonging to class 7.} 
We train an instance of \ac{IF} with 100 trees and $\psi=32$ on the \texttt{Glass} dataset. We obtain an F1-score equal to 0.7706. 
The model is able to correctly identify 28 out of 29 anomalies in class 7.

\subsubsection{Interpreting Isolation Forest Predictions: Comparison of \ourmethod  with KernelSHAP and LocalDIFFI on Glass dataset}
\begin{figure}
  \centering
  \begin{subfigure}{0.73\textwidth}
    \includegraphics[width=\linewidth]{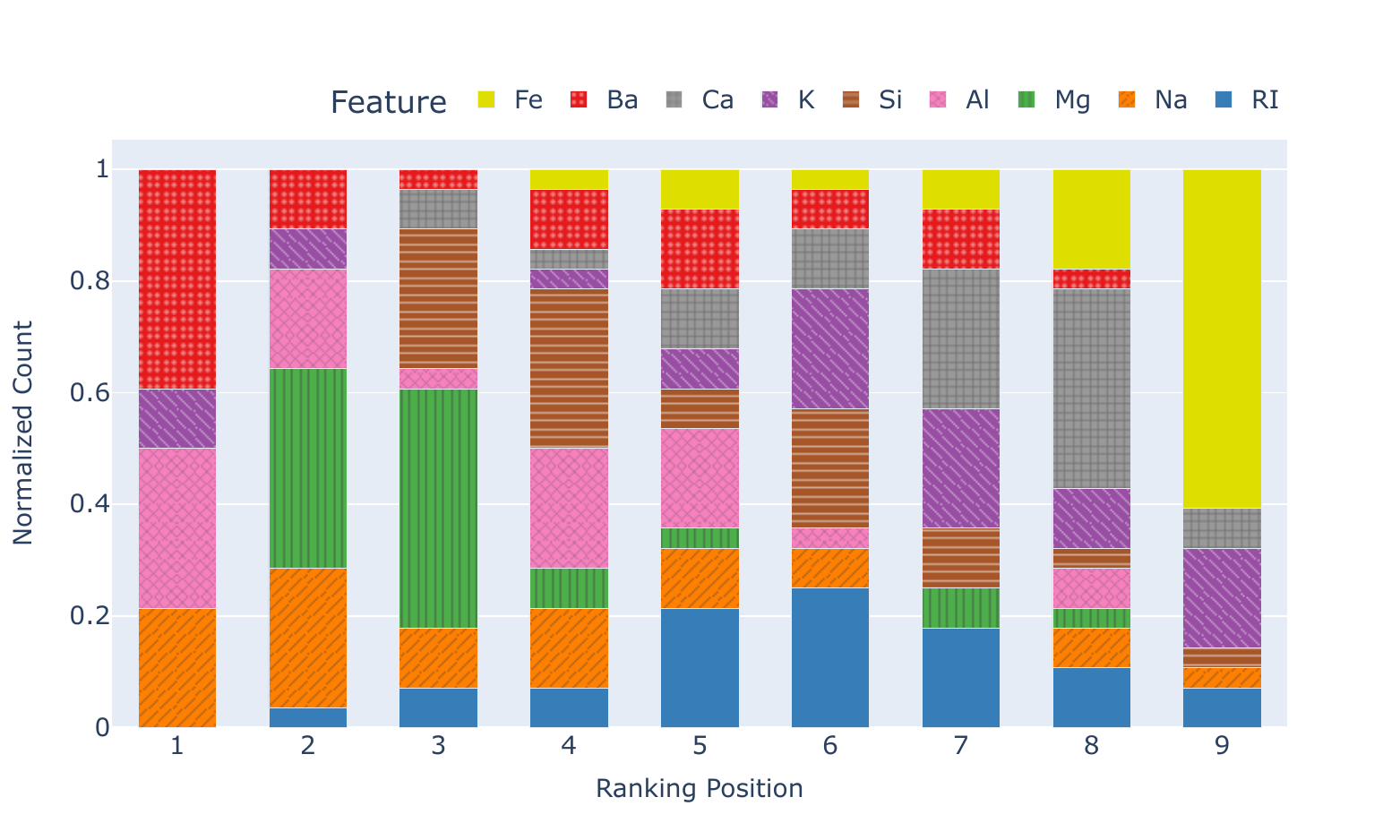}
    \caption{\ourmethod}
    \label{subfig:glass_acme_default_wights}
  \end{subfigure}
  \hfill
    \begin{subfigure}{0.73\textwidth}
    \includegraphics[width=\linewidth]{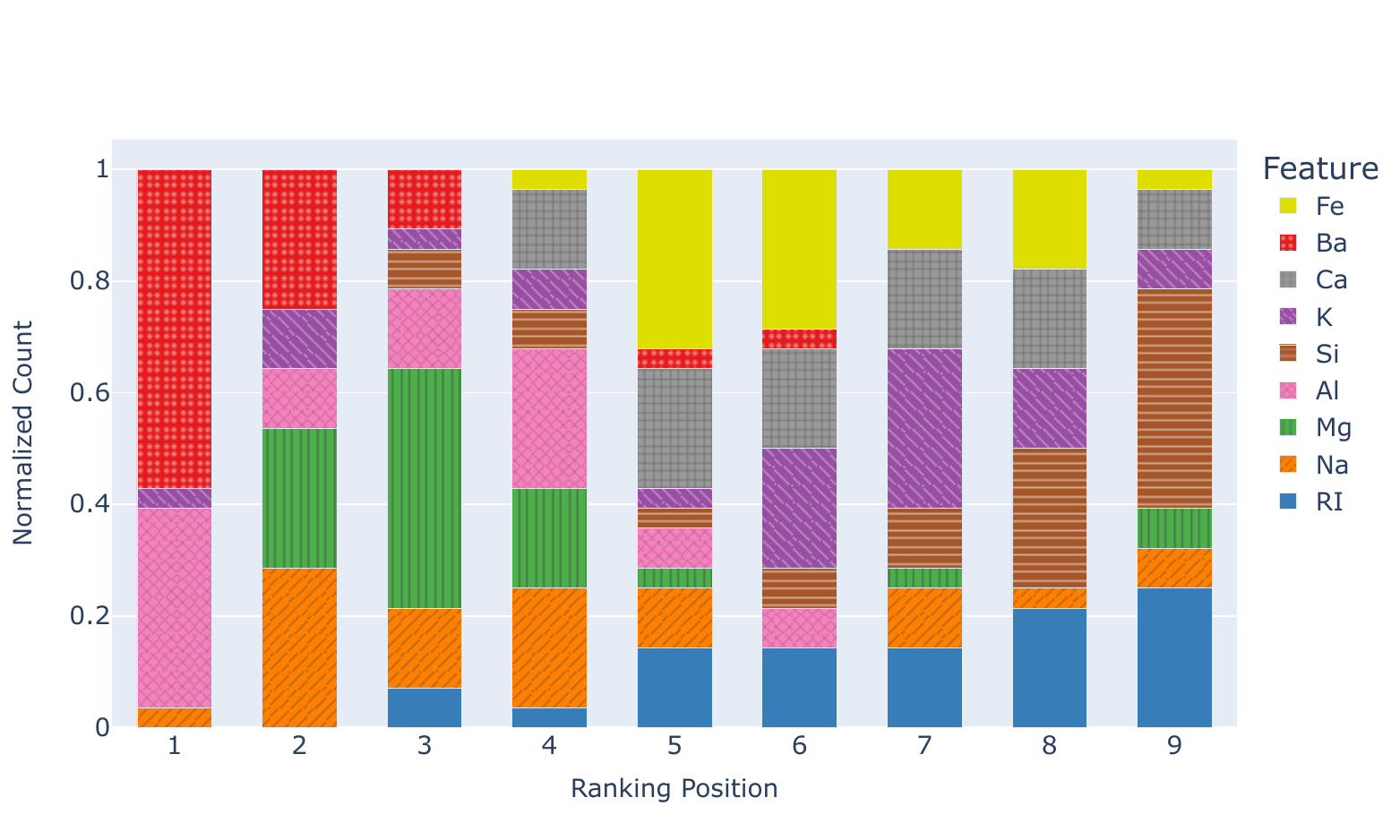}
    \caption{KernelSHAP}
    \label{subfig:glass_kernelshap}
  \end{subfigure}
  \hfill
  \begin{subfigure}{0.73\textwidth}
    \includegraphics[width=\linewidth]{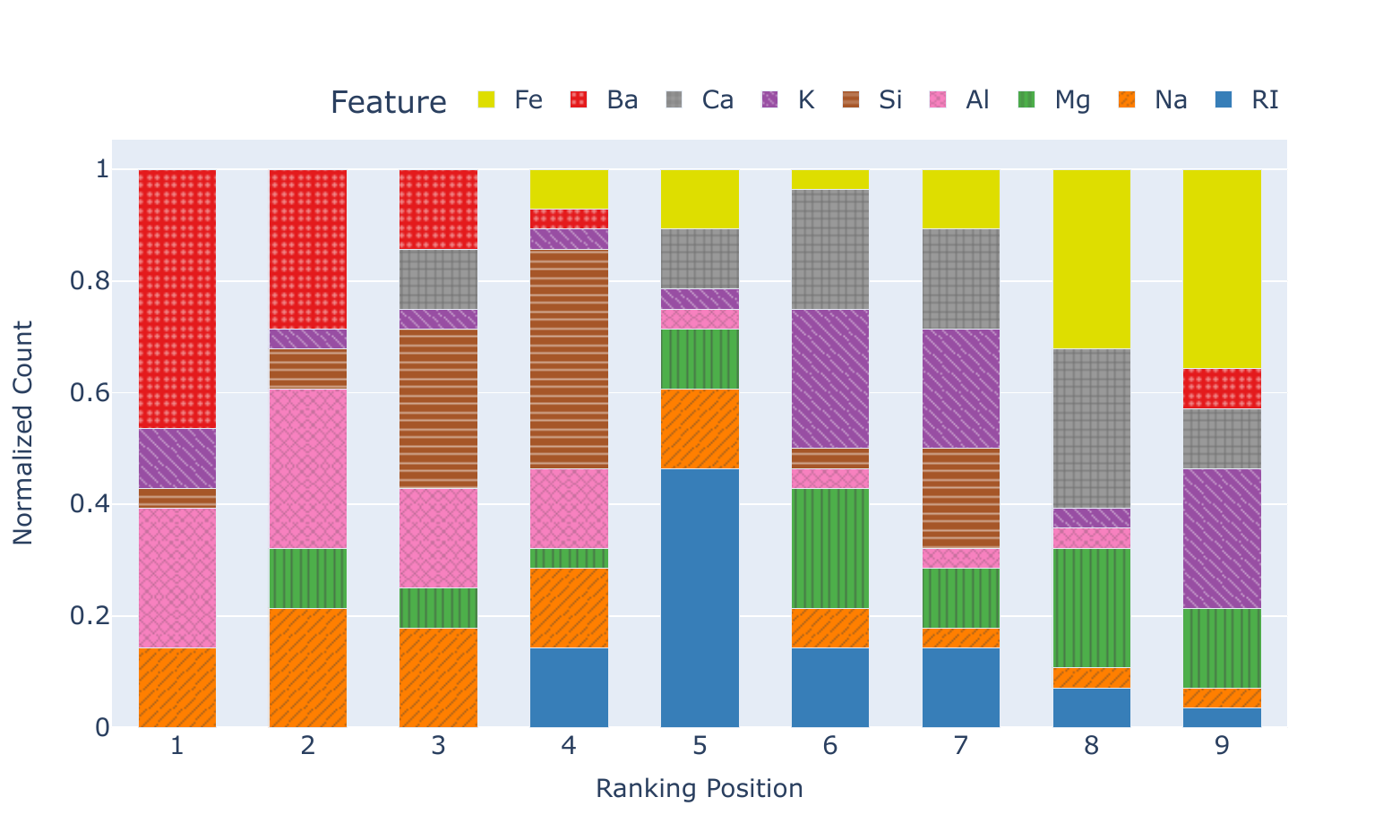}
    \caption{LocalDIFFI}
    \label{subfig:glass_diffi}
  \end{subfigure}

  \caption{Comparison of feature rankings computed by \ourmethod and KernelSHAP for an Isolation Forest trained on the Glass Dataset as detailed in \ref{subsec:realworlddatasets}.  Feature ranking based on \ourmethod with default weights in (a), KernelSHAP in (b) and LocalDIFFI in (c).}
  \label{fig:glass_acme_vs_diffi_vs_shap}
\end{figure}

We run \ourmethod with default weights to compute local explanations for anomalies detected within class 7. Fig. \ref{subfig:glass_acme_default_wights} illustrates the resulting feature ranking distributions across anomalies. As elaborated in Section \ref{subsec:syntheticdatasets}, the stacked bars at position $k$ represent the percentage of anomalous observations for which each feature is ranked at $k$-th position. 

We compare \ourmethod with KernelSHAP, which rankings are shown in \ref{subfig:glass_kernelshap}. Both approaches consistently identify \textsf{Ba} or \textsf{Al} as the most significant features across the majority of anomalies, aligning with prior knowledge. Additionally, the distributions are reasonably comparable. For example, \textsf{Mg} is ranked by both methods in the second, third or fourth position for the majority of data points. 
However, it is noteworthy that KernelSHAP computes the feature importance score only evaluating the impact of the feature perturbation on the anomaly score. On the other hand, \ourmethod provides more comprehensive information, by also enabling the possibility to assess which features contributed to change the classification of individual data points, as visualized in Fig. \ref{fig:loc_vis_example}.

Afterwards, we compare \ourmethod with LocalDIFFI \cite{carletti2023interpretable}, a model-specific feature importance approach for \ac{IF}. In Fig. \ref{subfig:glass_diffi} we present the feature rankings produced by LocalDIFFI. Once again, \textsf{Ba} and \textsf{Al} emerges as the most important features in most anomalies, as expected. Moreover, features like \textsf{Na} and \textsf{K} receive high rankings from both \ourmethod and LocalDIFFI in a non-negligible number of anomalies. Notably, the rankings of \textsf{Na} generated by both methods exhibit a similar pattern across other positions as well.

In summary, our analysis reveals that while there are slight variations arising from different computational mechanisms inherent in the three methods, the rankings of features proposed by each algorithm are qualitatively comparable, especially at the top positions. However, \ourmethod has potentially higher applicability. Indeed, it is agnostic, differently from LocalDIFFI and similarly to KernelSHAP. Compared to the latter, \ourmethod is much faster as detailed in in the end of this section.

\subsubsection{Analysis of \ourmethod Sub-scores on Glass dataset}
\begin{figure}
    \centering
    \includegraphics[width=\textwidth]{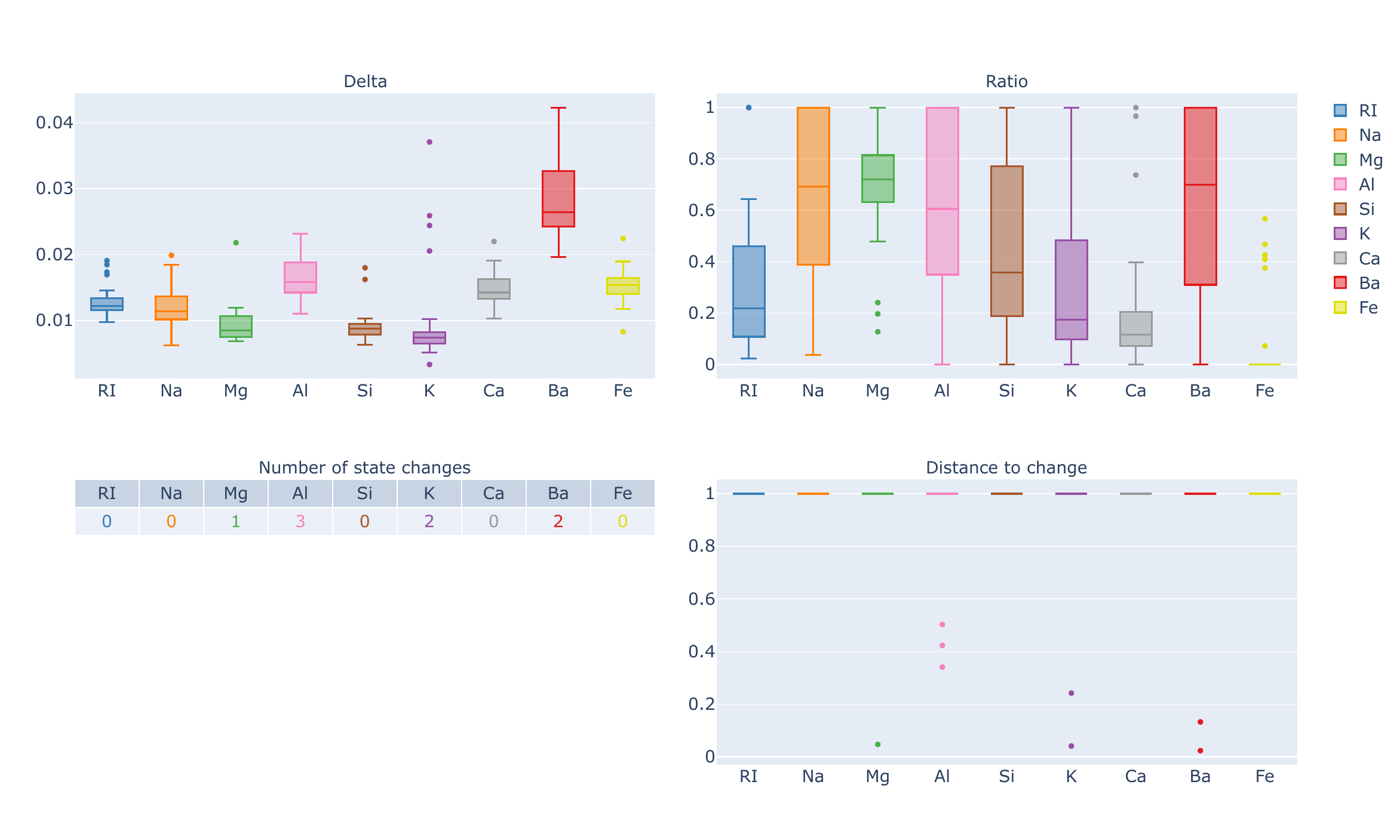}
    \caption{Distribution analysis of \ourmethod's sub-scores for predicted class 7 anomalies in the Glass dataset (Isolation Forest trained as detailed in \ref{subsec:realworlddatasets})}
    \label{fig:glass_IF_subscore_distribution}
\end{figure}

In this section we offer insights into the distributions of \ourmethod sub-scores Delta, Ratio, Change and Distance-to-change, introduced in Section \ref{subsec:localacmead}, that yielded the feature ranking displayed in \ref{subfig:glass_acme_default_wights}. Through boxplots visualized Fig. \ref{fig:glass_IF_subscore_distribution}, we provide a statistical description of these sub-scores across local explanations of anomalies in class 7 of the \texttt{Glass} dataset. The plot associated to Delta, situated at the top left, distinctly highlights that the feature \textsf{Ba} induces the most substantial perturbation of the anomaly score. On median, the second most relevant feature according to Delta is \textsf{Al}. Perturbations in \textsf{Al} lead to the highest number of class changes, observed in 3 anomalies, as detailed in the table at the bottom left of Fig. \ref{fig:glass_IF_subscore_distribution}. When solely considering the number of class changes, \textsf{Ba} and \textsf{K} would hold the second position in the ranking, with \textsf{Mg} obtaining the third. If we also take into account the Distance-to-change score, positioned at the bottom right, \textsf{Ba}, \textsf{K} and \textsf{Mg} emerges as the features that require the smallest perturbation to transition the corresponding anomalies to a normal state. Finally, the Ratio score, at the top right, does not distinctly reveal a consistent feature ranking across the observed anomalies.

\subsubsection{Computational time comparison between KernelSHAP and \ourmethod}

In this section, we compare the computational time required by KernelSHAP and \ourmethod using the \texttt{Satellite} dataset. For this experiment, we do not resort to \texttt{Glass} dataset due to its small size (213 data points, 51 of which anomalies) and the relatively low number of features ($p=9$), which wouldn't fully demonstrate the computational advantage of \ourmethod, which is instead very efficient with high-dimensional and large-size data.  
We train an \ac{IF} model with $\psi=256$ and 100 random trees, identifying 827 anomalies on 6435 data points, with a precision of 0.828, recall of 0.406 and an F1-Score of 0.545.

\begin{figure}
    \centering
    \includegraphics[width=\textwidth]{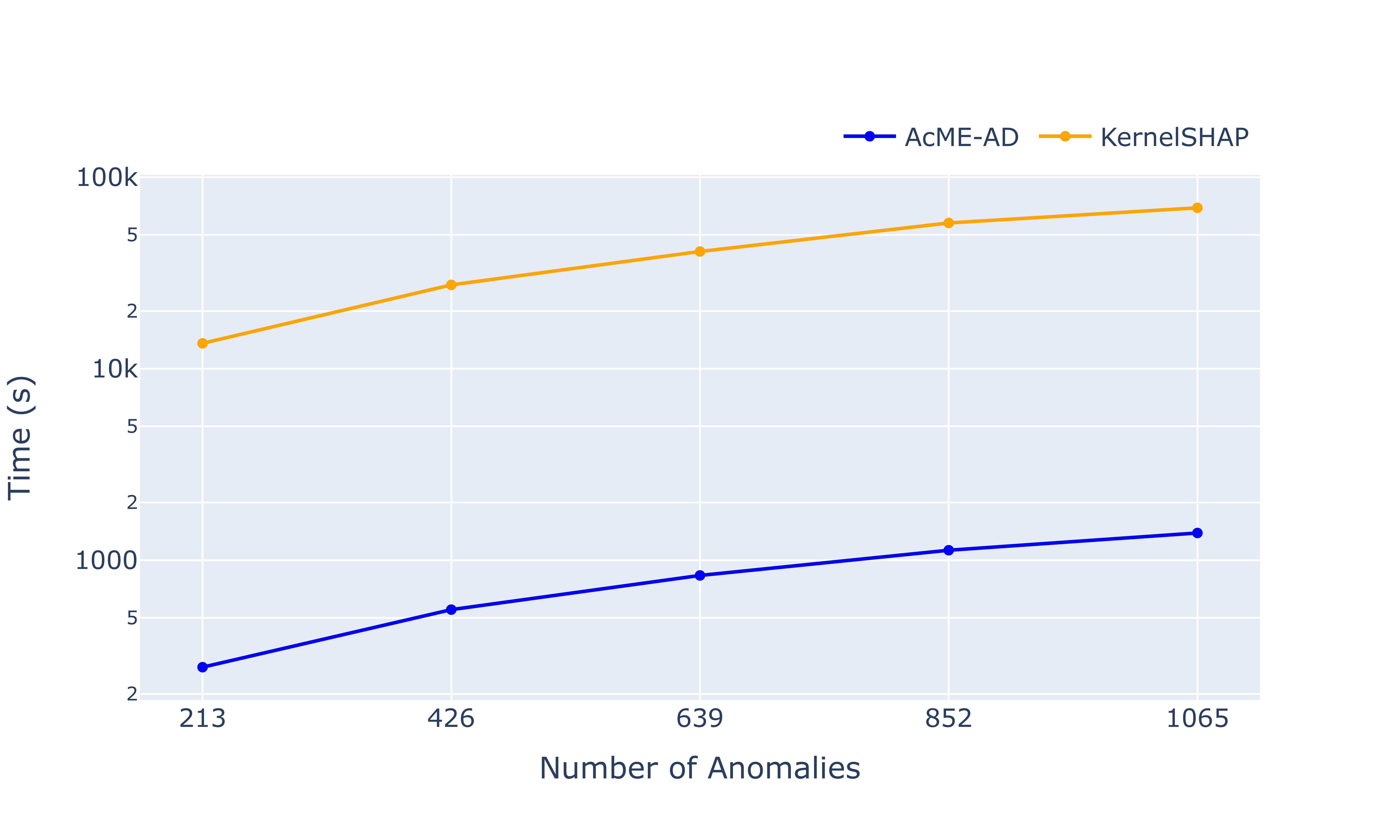}
    \caption{Comparison of computational time required by \ourmethod and KernelSHAP to produce local explanations for an Isolation Forest trained on the Satellite dataset as detailed in Section \ref{subsec:realworlddatasets}}
    \label{fig:time1}
\end{figure}

Figure \ref{fig:time1} compares the computational time\footnote{For reference, experiments are conducted on AMD Ryzen 7 3700X, 3.6GHz, RAM 20GB} of KernelSHAP and \ourmethod for explaining different number of anomalies, respectively the 20\%, 40\%, 60\%, 80\% and 100\% of the 827 predicted anomalies in the entire dataset. The fitting of both explainers was performed on the entire training set. The number of coalitions considered in KernelSHAP is the default value, i.e., $K=2p+2048$, where $p$ is the number of features. It is evident that \ourmethod outperforms KernelSHAP in terms of computational time. Notably, explaining 100\% of anomalies with \ourmethod is faster than explaining 20\% of anomalies with KernelSHAP. 
The high computational burden of KernelSHAP arises from two aspects. The first is the need to estimate a number of linear models that increase with the size of the dataset. The second is that estimating these linear models involves a matrix inversion which dimension increases with the number of sampled coalitions, which default value $K=2p+2048$, in turn, increases with the number of features $p$.

To cope with the required high number of linear models, the authors of \cite{APISHAP} suggest to reduce the size of the dataset used as background, needed to integrating out features, by sampling $\Tilde{N}$ rows from the original dataset.
Table \ref{tab:time_comparison_sample_dataset} shows the computational times required to explain ten anomalies using different size of the background dataset. It is important to underscore that, as demonstrated in \ac{AcME} \cite{dandolo2023acme}, sampling the dataset may lead to unreliable results. Nonetheless, \ourmethod exhibits one order of magnitude faster computation time even when utilizing KernelSHAP with a reduced background dataset size of 25\%.

\begin{table}
    \centering
    \small 
    \setlength{\tabcolsep}{12pt} 
    \begin{tabular}{lrr} 
        \hline
         & No. of background samples   & Elapsed time (s) \\
        \hline
        \ourmethod & 6435 (100\%) & 15.15 \\
        KernelSHAP & 1609 (25\%) & 171.79 \\
        KernelSHAP & 3218 (50\%) & 358.83 \\
        KernelSHAP & 4826 (75\%) & 544.12 \\
        KernelSHAP & 6435 (100\%) & 685.53 \\
        \hline
    \end{tabular}
    \vspace{10pt}
    \caption{Elapsed Time to explain 10 data points for \ourmethod with complete dataset and for KernelSHAP with different numbers of background samples}
    \label{tab:time_comparison_sample_dataset}
\end{table}
To address the matrix inversion issues, the only viable option is to reduce the number of coalitions $K$, even if, again, it can be detrimental, especially when the feature distribution is nontrivial \cite{dandolo2023acme}. Table \ref{tab:time_comparison_n_coalitions} displays the times required to generate explanations for ten anomalies, with complete background dataset, across the 25\% 50\%, 75\% and 100\% of $2p+2048$ number of coalitions. 

\begin{table}
    \centering
    \small 
    \setlength{\tabcolsep}{12pt} 
    \begin{tabular}{lrr} 
        \hline
         & No. of coalitions  & Elapsed time (s) \\
        \hline
        \ourmethod & - & 15.15 \\
        KernelSHAP & 530 (25\%) & 197.10 \\
        KernelSHAP & 1060 (50\%) & 345.05 \\
        KernelSHAP & 1590 (75\%) & 491.57 \\
        KernelSHAP & 2120 (100\%) & 644.44 \\
        \hline
    \end{tabular}
    \vspace{10pt}
    \caption{Elapsed Time to explain 10 data points for \ourmethod and for KernelSHAP with different numbers of coalitions}
    \label{tab:time_comparison_n_coalitions}
\end{table}

As these experiments highlight, \ourmethod demonstrates its capability in time-sensitive environments. Thus, it turns out to be a valuable tool for scenarios where rapid interpretability is critical, compared to methods with potentially higher computational demands.

\subsection{Feature Selection}
\label{subsec:featureselection}
To further assess the robustness of \ourmethod, we employ feature selection as a proxy task. 
We evaluate the F1-Score of the \ac{AD} model trained by restricting the input features to the most relevant according to \ourmethod. We compare these performance metrics with a benchmark random feature selection.
As highlighted in \cite{wang2023feature}, numerous real-world datasets for \ac{AD} have a large proportion of noisy, redundant or irrelevant features. These types of features and their complex interaction can result in high computational cost and a negative effect in the \ac{AD} algorithms \cite{yang2019isolation}. Therefore, we expect that removing these features, which should coincide with the least important features according to \ourmethod, should enhance the algorithm performance.

To perform feature selection, the importance of features must be evaluated globally, on the entire model. In order to achieve a global evaluation we perform the following steps. We train $N_{f_s}=5$ instances of the \ac{AD} model on the same dataset but with different seeds. For each model instance, we compute the global importance score for each feature by summing the corresponding local importance scores computed on predicted anomalies (with default weights for the sub-scores). Then, we aggregate the global scores to define a feature ranking as follows:  
\begin{enumerate}
    \item Define $S_{agg}\in \mathbb{R}^p$, initialized with zeros, where $p$ is the number of features; 
    \item For each one of the $N_{f_s}$ instances of the model: 
    \begin{itemize}
        \item we sort the Global Importance Scores in descending order, obtaining a ranking of feature from the most to the last important feature for the specific instance; 
        \item update $S_{agg}$ by summing, for each feature $j$, the quantity $ \Delta_{f_s,j} = 1 - \log(\hat{r}_j)/{\log(p)}$. This value depends on the rank position $\hat{r}_j$ of the feature $j$ and it is higher the higher the position in the ranking.
    \end{itemize}
    \item $S_{agg}$ is used to define the ranking of features, the higher the score the more important the feature. 
\end{enumerate}

After computing the aggregated score vector $S_{agg}$, we perform feature selection. Specifically, we iterate through values of $k$ from 1 to the total number of feature. At each step $k$, we train 50 instances of \ac{IF} with different random seeds using only the top $k$ most important features according to $S_{agg}$. We compute the median F1-scores for each $k$. 
Similarly, we conduct experiments by selecting $k$ random features and compare the resulting median F1-scores. 

\begin{figure}[h]
    \centering
    \includegraphics[width=\textwidth]{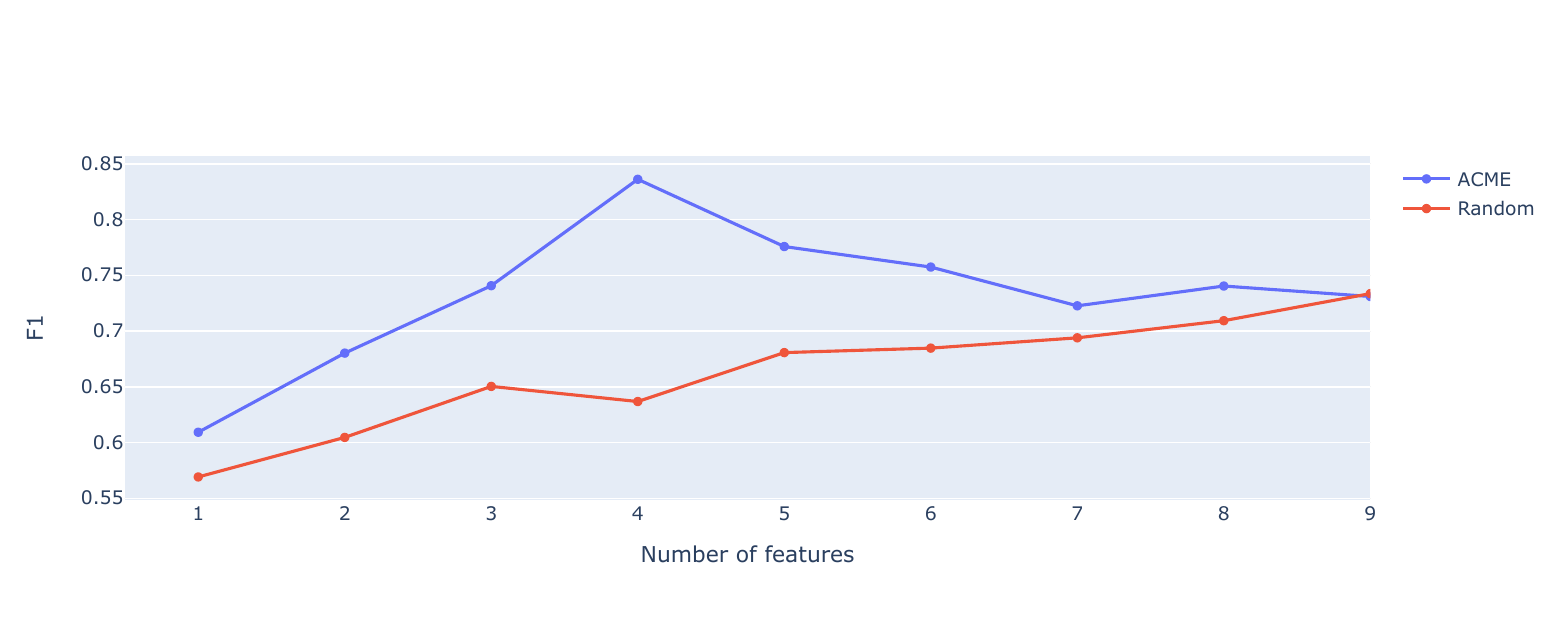}
    \caption{F1 scores obtained on the Glass dataset through feature selection, as detailed in Section \ref{subsec:featureselection}}
    \label{fig:fs_glass}
\end{figure}

\begin{figure}[h]
    \centering
    \includegraphics[width=\textwidth]{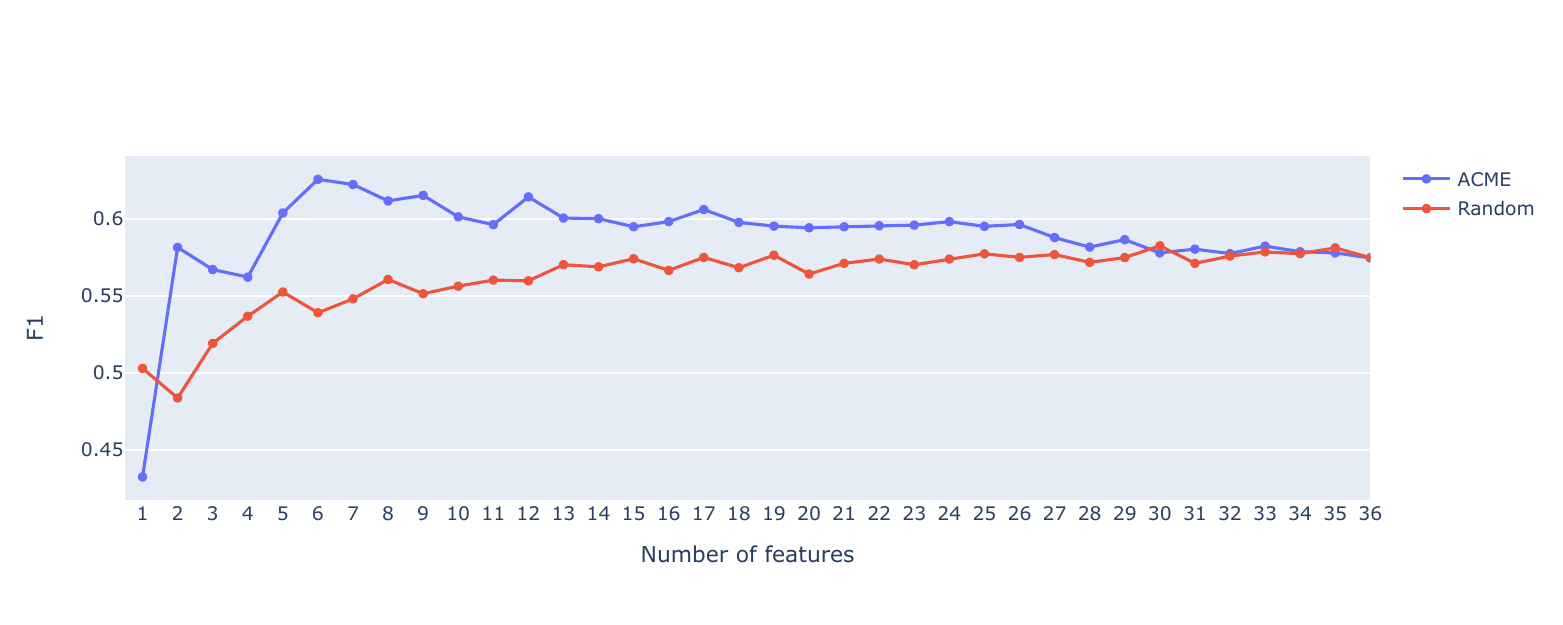}
    \caption{F1 scores obtained on the Satellite dataset through feature selection, as detailed in Section \ref{subsec:featureselection}}
    \label{fig:fs_satellite}
\end{figure}

Fig. \ref{fig:fs_glass} and Fig. \ref{fig:fs_satellite} plot the median F1-scores for the \texttt{Glass} and the \texttt{Satellite} dataset respectively. \ourmethod (the blue line) achieves higher F1-scores compared to random feature selection, demonstrating that the selected features are indeed the most relevant for the model and for the task (the only exception is represented for \texttt{Satellite} when selecting only one feature \ref{fig:fs_satellite}). 
For \texttt{Glass} dataset, our method identify four as the optimal number of features (which are \textsf{Mg, Ca, Na, RI}), while for \texttt{Satellite}, it specifies six features (which are \textsf{7, 5, 18, 3, 33, 29}). Note that these features are indicated as the most relevant to explain the entire set of outliers and, particularly for \texttt{Glass}, not only class 7 as done in previous experiments.

\section{Future Works and Conclusions}
\label{sec:conclusions}

This work introduces \ourmethod, a model-agnostic framework for interpreting anomaly detection predictions in the unsupervised learning scenario. By efficiently enabling a deeper understanding of how anomalies are identified, the proposed approach enhances the trust and usability of anomaly detection systems. Experimental results suggest it to be particularly suitable for scenarios where rapid interpretability is critical.

Most \ac{AD} algorithms both compute an anomaly score for each data point and classify samples as either anomalous or normal. However, standard \emph{post hoc} interpretability approaches fail to offer insights into both anomaly scores and the predicted class, limiting the understanding of anomalous behavior. 
Instead, \ourmethod shows the features roles on both the anomaly score and on the predicted class (normal or anomaly). It actually goes beyond this by allowing the user to decide how to weight these two aspects, depending on the application. To the best of our knowledge, it is the first approach with this characteristic. 

Our future research will focus on several key areas to refine and extend our interpretability framework:
\begin{itemize}
\item Feature Interactions: A deeper exploration into how features interact within models will be pursued. Understanding these interactions is crucial for providing more detailed and accurate interpretations, especially in complex datasets where the relationships between variables significantly impact model outcomes.
\item Visualization Refinement: Based on real-world application feedback, we plan to improve our visualization tools. Effective visualizations are essential for conveying complex model behaviors in an accessible manner, and user feedback will be invaluable for making these tools more intuitive and informative.
\item Active Learning Integration: Incorporating Active Learning principles to use interpretability as a feedback loop for model refinement is another research direction we aim to explore. By aligning the model more closely with expert domain knowledge through feedback, we anticipate not only improved model performance but also increased adoption of machine learning solutions in practice.
\item Reducing Misclassifications through Interpetability: Interpretability will also be used as a diagnostic tool to address and reduce false positives and negatives. By understanding the reasons behind these errors, targeted improvements can be made to enhance the overall accuracy and reliability of anomaly detection models.
\end{itemize}
By addressing these future research directions, we aim to push the boundaries of interpretability in unsupervised anomaly detection, ultimately creating more reliable and impactful solutions for the real world.

\bibliographystyle{splncs04}
\bibliography{references.bib}
%




\end{document}